\documentclass[lettersize,journal]{IEEEtran}
\usepackage{amsmath,amsfonts}
\usepackage{algorithmic}
\usepackage{algorithm}
\usepackage{array}
\usepackage[caption=false,font=normalsize,labelfont=sf,textfont=sf]{subfig}
\usepackage{textcomp}
\usepackage{stfloats}
\usepackage{float}
\usepackage{url}
\usepackage{verbatim}
\usepackage{graphicx}
\usepackage{cite}
\usepackage[table]{xcolor}
\usepackage{xcolor}
\usepackage{amsmath,amssymb}
\usepackage{booktabs}
\usepackage{multirow}
\usepackage{adjustbox}
\usepackage{tabularx}

\usepackage{colortbl}
\usepackage{makecell}
\newcolumntype{C}{>{\centering\arraybackslash}X}
\definecolor{oursrow}{RGB}{237,244,250}

\usepackage{hyperref}
\usepackage{orcidlink}
\hypersetup{
    colorlinks=true,
    linkcolor=blue!55!black,  
    citecolor=blue!55!black,  
    urlcolor=blue!55!black    
}

\newcolumntype{Y}{>{\centering\arraybackslash}X}

\begin{document}

\title{Reduce the Artifact Bias for More Generalizable AI-Generated Image Detection}


\author{
	Yiheng Li\,\orcidlink{0000-0002-1182-1176}\IEEEauthorrefmark{1},
    Yang Yang\,\orcidlink{0000-0003-0559-5464}\IEEEauthorrefmark{1}\IEEEauthorrefmark{2},~\IEEEmembership{Member,~IEEE},
	Wenhao Wang\,\orcidlink{0000-0001-8727-1572}\IEEEauthorrefmark{1}\IEEEauthorrefmark{2}, \\
	Zichang Tan\,\orcidlink{0000-0002-8501-4123},~\IEEEmembership{Member,~IEEE},
    Zecheng Lin\,\orcidlink{0009-0008-2609-6713},
    Li Gao\,\orcidlink{0009-0005-3726-248X},
    Zhen Lei\,\orcidlink{0000-0002-0791-189X},~\IEEEmembership{Fellow,~IEEE}
\thanks{\IEEEauthorrefmark{1}Yiheng Li, Yang Yang, and Wenhao Wang contributed equally to this work. \IEEEauthorrefmark{2}Yang Yang and Wenhao Wang are co-corresponding authors.}
\thanks{Yiheng Li and Yang Yang are with the School of Artificial Intelligence, University of Chinese Academy of Sciences, Beijing 100049, China; the State Key Laboratory of Multimodal Artificial Intelligence Systems, Institute of Automation, Chinese Academy of Sciences, Beijing 100190, China. (e-mail: liyiheng2024@ia.ac.cn, yangyang2013@ia.ac.cn)}
\thanks{Wenhao Wang is with the University of Technology Sydney, Sydney, Australia. (e-mail: wangwenhao@vastilab.com)}
\thanks{Zichang Tan is with the Sangfor Technologies Inc. (e-mail: tanzichang@foxmail.com)}
\thanks{Zecheng Lin is with Tsinghua University, Beijing 100084, China. (e-mail: zc-lin24@mails.tsinghua.edu.cn)}
\thanks{Li Gao is with the China Mobile Financial Technology Co., Ltd. (e-mail: gaolids@chinamobile.com)}
\thanks{Zhen Lei is with the State Key Laboratory of Multimodal Artificial Intelligence Systems, Institute of Automation, Chinese Academy of Sciences, Beijing 100190, China; the School of Artificial Intelligence, University of Chinese Academy of Sciences, Beijing 100049, China; the Centre for Artificial Intelligence and Robotics, Hong Kong Institute of Science and Innovation, Chinese Academy of Sciences, Hong Kong, China; the School of Computer Science and Engineering, the Faculty of Innovation Engineering, Macau University of Science and Technology, Macau, China.  (e-mail: zhen.lei@ia.ac.cn)}
}

\markboth{Journal of \LaTeX\ Class Files,~Vol.~14, No.~8, August~2021}%
{Shell \MakeLowercase{\textit{et al.}}: A Sample Article Using IEEEtran.cls for IEEE Journals}


\maketitle

\begin{abstract}
As the misuse of AI-generated images grows, generalizable image detection techniques are urgently needed. Recent state-of-the-art (SOTA) methods adopt aligned training datasets to reduce content, size, and format biases, empowering models to capture robust forgery cues. A common strategy employs reconstruction techniques, e.g., VAE and DDIM, to construct aligned synthetic negatives. However, relying on a single reconstruction process yields a narrow and homogeneous artifact distribution, leaving forensic traces from other artifact-forming mechanisms underrepresented. To broaden artifact coverage without sacrificing alignment, we construct adversarial restoration-based negatives with SRGAN, whose learned upsampling and texture restoration yield traces complementary to VAE reconstruction while preserving content, size, and format. Directly mixing the two aligned fake domains is nontrivial because their artifact manifolds and optimization directions can conflict. We therefore propose Artifact-Complementary Expert Fusion (ACEF), a two-stage framework for robust AIGC detection. ACEF first constructs two artifact-specific experts via LoRA adaptation on a frozen foundation backbone. It then freezes them and introduces Layer-wise Artifact-Complementary Fusion (LACF) to integrate multi-layer source-specific and cross-artifact evidence through an adaptive gate. Extensive experiments on 13 diverse benchmarks demonstrate the effectiveness of our method.

\end{abstract}


\section{Introduction}
Generative AI \cite{ho2020denoising,goodfellow2014generative,van2016pixel} advances rapidly, enabling easy creation of hyper-realistic images for creative design, media production, scientific visualization, and the entertainment industry. However, malicious use of such technology can spread disinformation, facilitate fraud, and manipulate public opinion, undermining information authenticity and social trust. Consequently, reliable AI-generated image detection \cite{tan2023learning,tan2024rethinking,liu2024forgery,tan2024frequency} is urgently needed to mitigate these harms and enable responsible AIGC use. A critical challenge is achieving strong generalization \cite{ojha2023towards,li2025towards}, namely the ability to detect images generated not only by known models but also by unseen generators.

\begin{figure}[t]
    \centering
    \includegraphics[width=0.95\columnwidth]{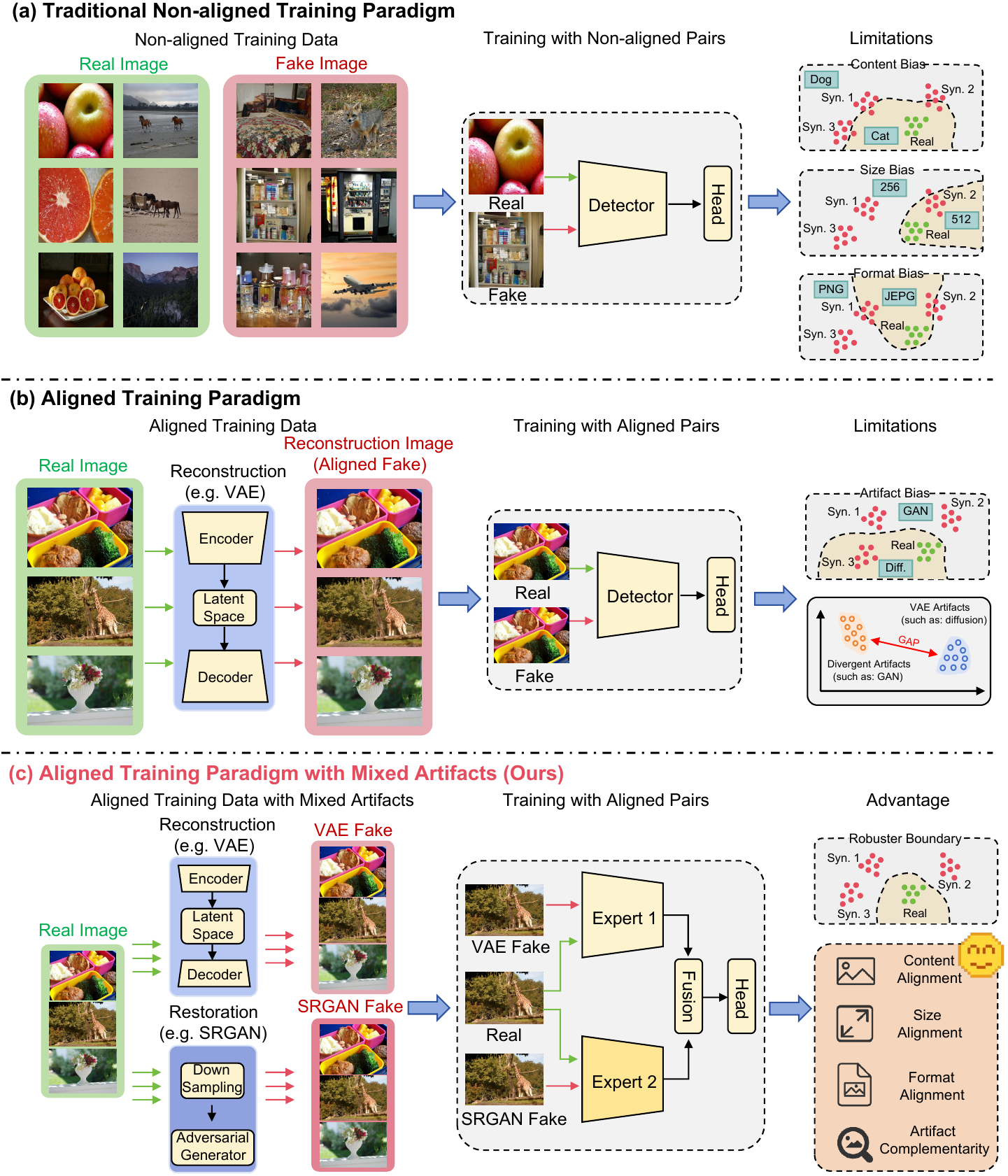}
    \caption{
    \textbf{Comparison of training paradigms for AI-generated image detection.} (a) Conventional non-aligned training suffers from content, size, and format biases. (b) Reconstruction-based aligned training reduces these biases by generating VAE/DDIM-style aligned fakes, but its artifacts remain concentrated around reconstruction-induced traces. (c) We complement reconstruction-based negatives with adversarial restoration-based aligned negatives generated by SRGAN, and use ACEF to combine heterogeneous artifact cues without direct mixed-training interference.}
    \label{fig1}
\end{figure} 

Early training protocols for AI-generated image detection often collect real and fake images from different sources and directly train a binary classifier, as illustrated in Fig.~\ref{fig1}~(a). Under this paradigm, existing methods mainly improve generalization by learning transferable forensic representations, ranging from low-level pixel and frequency traces \cite{wang2020cnn,tan2024rethinking,tan2024frequency} to high-level features from pretrained visual models \cite{ojha2023towards,liu2024forgery}. However, these methods remain vulnerable to shortcuts caused by source-dependent differences between real and fake training images, which can overshadow intrinsic synthesis traces and weaken cross-domain generalization.

To reduce these shortcuts, recent methods adopt an aligned training paradigm \cite{rajan2024aligned,chen2024drct,chen2025dual,guillaro2025bias,liu2025beyond}, which constructs synthetic negatives from real-image anchors to control content, size, and format differences. As illustrated in Fig.~\ref{fig1}~(b), comparing each real image with its aligned synthetic counterpart suppresses source-dependent cues and encourages detectors to focus on artifacts introduced during reconstruction. Specifically, DRCT \cite{chen2024drct} constructs challenging negatives through diffusion reconstruction, AlignedForensics \cite{rajan2024aligned} uses VAE reconstruction to align real/fake content and size, and DDA \cite{chen2025dual} further reduces pixel- and frequency-level discrepancies.

However, aligned real/fake correspondence does not ensure that supervision spans diverse artifact-forming mechanisms. Most existing methods construct aligned synthetic negatives with a single reconstruction process, such as VAE \cite{rombach2022high,chen2025dual,rajan2024aligned} or DDIM \cite{song2020denoising,chen2024drct}. Although these operators differ, each training set is dominated by artifacts from its selected pipeline, including latent compression, decoder reconstruction, or denoising dynamics. Mechanistically distinct traces from adversarial texture synthesis, learned upsampling, local structural distortion, and spectral mismatch therefore remain underrepresented \cite{goodfellow2014generative,odena2016deconvolution,durall2020watch,tan2024rethinking}. Consequently, a detector may still form its decision boundary around only a narrow portion of the broader forgery artifact space. We term this remaining limitation \textbf{artifact bias}. Addressing it requires answering two coupled questions: how to construct complementary aligned negatives with broader artifact coverage, and how to learn from heterogeneous artifacts without negative transfer.

\textbf{How to construct an aligned dataset with broader artifact coverage?}
An appropriate complementary source must preserve real-anchor alignment while introducing complementary forensic traces beyond those typically induced by standard VAE reconstruction. SRGAN \cite{ledig2017photo} provides such a source through a deterministic degradation-restoration pipeline that retains semantic content, structure, size, and format, while adversarial supervision and learned upsampling introduce distinct texture, structural, and frequency traces. We use SRGAN as a controlled proxy for an alternative artifact-forming mechanism rather than the entire GAN family. Together, VAE reconstruction and SRGAN restoration construct two aligned fake domains from shared real anchors through distinct artifact-forming processes, as illustrated in Fig.~\ref{fig1}~(c).

\textbf{How to train on mixed aligned artifacts without negative transfer?}
The resulting training set contains heterogeneous fake domains with shared real anchors. Directly mixing them in one detector can be problematic because our empirical and gradient analyses reveal incompatible feature manifolds and optimization directions. We therefore propose Artifact-Complementary Expert Fusion (ACEF), which trains artifact-specific experts independently and combines them through lightweight layer-wise fusion, preserving specialized forensic knowledge while learning a unified decision boundary.

Across cross-domain GAN and diffusion generators, multiple benchmarks, localized forgeries, and robustness settings, ACEF demonstrates strong and balanced generalization. Our contributions are threefold:
\begin{itemize}
    \item \textbf{Identification of artifact bias.} We identify incomplete artifact coverage as a remaining limitation of aligned training and analyze its impact on model generalization.
    \item \textbf{Artifact-complementary aligned data construction.} We introduce SRGAN-based adversarial restoration negatives that complement VAE reconstruction while preserving content, size, and format alignment.
    \item \textbf{Artifact-complementary expert fusion.} We propose ACEF to preserve source-specific expertise and integrate heterogeneous aligned artifacts without direct mixed-training interference.
\end{itemize}

\section{Related Work}

\subsection{Artifact-Based AI-Generated Image Detection}
Early detectors exploit traces left by image synthesis pipelines. CNNSpot \cite{wang2020cnn} shows that a classifier trained on one generator can transfer to unseen generators, indicating that synthetic images share detectable low-level statistics. Subsequent methods refine these cues across spatial, color, and frequency domains. NPR \cite{tan2024rethinking} connects neighboring-pixel correlations to upsampling operations, FerretNet \cite{liang2025ferretnet} models local pixel dependencies, and FALCON-Net \cite{zhang2026falcon} aggregates sensor-noise and local-variation patterns. Color-distribution analysis \cite{jia2025secret} instead captures chromatic deviations introduced by synthesis pipelines. Frequency-based methods complement these spatial cues by characterizing spectral mismatch. Frequency-aware learning \cite{tan2024frequency} emphasizes weak spectral traces that are less visible in RGB space. Dual-frequency modeling \cite{yan2026dual} combines complementary frequency bands, SAFE \cite{li2025improving} exposes architectural traces through high-frequency transformations, and noise-informed attention \cite{guan2025noise} captures anomalous noise patterns shared across diffusion models.

Another branch detects synthetic images through their reconstruction or inversion behavior. SeDID \cite{ma2023exposing} exploits errors accumulated during deterministic diffusion inversion and denoising, whereas DIRE \cite{wang2023dire} uses discrepancies between an input and its diffusion reconstruction. FakeInversion \cite{cazenavette2024fakeinversion} extracts inversion and reconstruction features from Stable Diffusion to detect unseen text-to-image models. LaRE$^2$ \cite{luo2024lare} models reconstruction evidence in latent space, while FIRE \cite{chu2025fire} strengthens reconstruction errors with frequency guidance. These methods reveal informative model-response cues, although their generalization can remain dependent on the selected auxiliary model and reconstruction process.

\subsection{AI-Generated Image Detection via Foundational Model}
Visual foundation models provide transferable representations learned from large-scale pre-training. Most detectors in this line build on CLIP \cite{radford2021learning}, whose vision-language pre-training provides both semantic knowledge and reusable visual features. UnivFD \cite{ojha2023towards} demonstrates that a linear classifier on frozen CLIP features can generalize across generator families. Later CLIP-based methods adapt prompts, intermediate representations, or auxiliary forensic cues. FatFormer \cite{liu2024forgery} introduces forgery-aware adapters that combine vision-language representations with frequency evidence, while C2P-CLIP \cite{tan2025c2p} uses category-common prompts to reduce semantic overfitting. IAPL \cite{li2025towards} further conditions prompts on individual inputs, and Effort \cite{yan2024orthogonal} separates domain-specific artifacts from domain-invariant features through orthogonal subspace decomposition. RINE \cite{koutlis2024leveraging} maps features from multiple intermediate CLIP encoder blocks into a forgery-aware space and learns their relative importance. CLIP-ADA \cite{deng2026clipada} similarly aggregates multi-stage features to learn artifact-invariant representations, whereas Co-SPY \cite{cheng2025co} and AIDE \cite{yan2024sanity} combine CLIP semantics with complementary pixel- or frequency-level cues. GAPL \cite{qin2025scaling} addresses scaling across many generator sources by organizing heterogeneous evidence with generator-aware prototypes and a LoRA-adapted CLIP encoder. Beyond CLIP-based vision-language pre-training, self-supervised DINO \cite{caron2021emerging} backbones offer dense patch representations that preserve local texture and structural evidence. All Patches Matter \cite{yang2025all} addresses few-patch bias through randomized patch reconstruction and patch-wise contrastive learning, whereas Simplicity Prevails \cite{zhou2026simplicity} demonstrates the generalization benefit of strong pretrained backbones with simple detection heads. These methods improve representation and evidence fusion, but do not determine how heterogeneous aligned negatives should be constructed and learned together.

\begin{figure*}[t]
    \centering
    \includegraphics[width=0.95\textwidth]{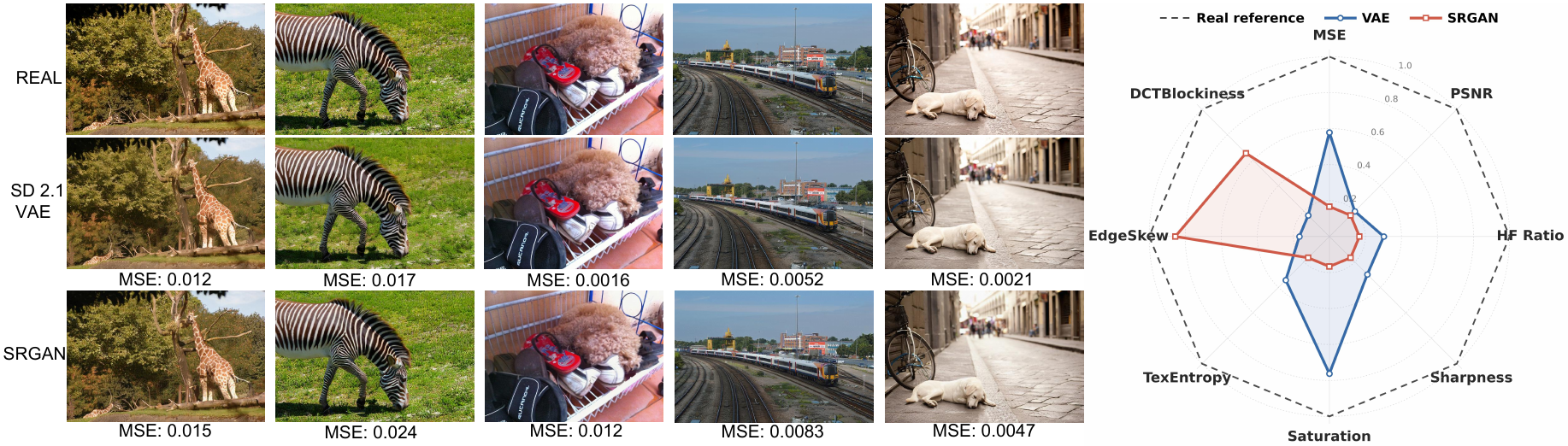}
    \caption{
    \textbf{Complementary artifact profiles produced by SD 2.1 VAE and SRGAN.} The radar chart reports real-referenced proximity scores in [0, 1] across eight metrics, where 1.0 (gray dashed line) indicates perfect alignment with real-image statistics. For each metric, the score is computed as $s_m=1-\lvert v_m-r_m\rvert/(\gamma d_m)$, where $v_m$ is the measured value, $r_m$ is the real-image reference, $d_m=\max(\lvert v_m^{\mathrm{VAE}}-r_m\rvert,\lvert v_m^{\mathrm{SRGAN}}-r_m\rvert)$, and $\gamma=1.2$. The metrics are computed from 1,000 randomly sampled images.}
    \label{fig2}
\end{figure*}

\subsection{Bias Reduction and Aligned Training Paradigm}
Beyond detector architecture, dataset construction strongly affects generalization. When real and fake images are collected through different pipelines, acquisition properties can become label-correlated shortcuts unrelated to intrinsic fakeness. These shortcuts include semantic content, image size or aspect ratio, compression format, source-specific preprocessing, collection procedures, and class imbalance. De-fake \cite{sha2023fake} identifies characteristic dimensions in text-to-image datasets, making image size predictive of the label. Fake or JPEG \cite{grommelt2024fake} shows that compression and format differences can dominate performance on common benchmarks. Semantic and category bias further motivates content-agnostic, category-aware adaptation to emerging generators \cite{tang2025extensible}. Consequently, high benchmark accuracy may reflect dataset separation rather than transferable generator traces, motivating explicit real/fake alignment.

Aligned training reduces these shortcuts by constructing synthetic negatives from real-image anchors. SemGIR \cite{yu2024semgir} uses semantic-guided regeneration for joint detection and attribution. DRCT \cite{chen2024drct} combines diffusion reconstruction with contrastive learning to produce difficult negatives, while B-Free \cite{guillaro2025bias} employs self-conditioned inpainting reconstruction and content augmentation. AlignedForensics \cite{rajan2024aligned} uses VAE reconstruction to align real and fake content and size. DDA \cite{chen2025dual} further reduces pixel- and frequency-level discrepancies, and REM \cite{liu2025beyond} forms real-centric reconstruction envelopes by injecting controlled latent noise. These methods suppress content, size, and format shortcuts, but reliance on a single reconstruction source limits artifact coverage across diverse generators. We instead study complementary aligned artifact sources under the same real-anchor constraint.

\section{Methods}

\subsection{Artifact Bias}
\label{sec:artifact-bias}

\begin{figure*}[t]
    \centering
    \includegraphics[width=0.95\textwidth]{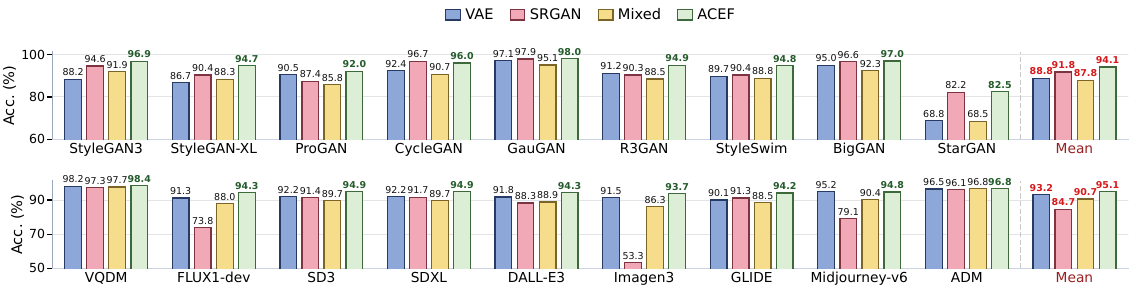}
    \caption{
    \textbf{Performance comparison of different training paradigms on GAN-based (Top) and diffusion-based (Bottom) benchmarks.} All variants use DINO-v3-L with LoRA ($r=8$, $\alpha=1$).}
    \label{fig3}
\end{figure*} 

\textbf{Reconstruction-Induced Bias.} Following prior aligned-training methods \cite{rajan2024aligned,chen2025dual}, we construct reconstruction-based negatives from the same real-image anchors. Specifically, the SD2.1 VAE \cite{rombach2022high} maps each real image $x \in \mathcal{D}_R$ to $x_{vae}=F_{vae}(x)$, yielding the paired training set $T_{vae}=\{(x,0),(F_{vae}(x),1)\mid x\in\mathcal{D}_R\}$. This pairing controls semantic content, spatial layout, size, and format, but fixes the artifact-forming operator to one VAE encoder-decoder. Its lossy bottleneck preferentially alters high-frequency details, local textures, and edges, encouraging a detector trained on $T_{vae}$ to fit decoder-specific traces while missing other synthesis processes. This incomplete coverage induces \textbf{artifact bias}.

\textbf{Complementary Artifact Construction via SRGAN.} We construct a complementary aligned domain with SRGAN \cite{ledig2017photo}. Given an anchor $x$, deterministic degradation $D(\cdot)$ and restoration produce $x_{srgan}=G_{srgan}(D(x))$, yielding $T_{srgan}=\{(x,0),(G_{srgan}(D(x)),1)\mid x\in\mathcal{D}_R\}$. Adversarial restoration and hierarchical upsampling introduce texture hallucination, local structural inconsistency, and spectral artifacts while retaining shared real anchors. Fig.~\ref{fig2} shows that the two domains deviate from real-image statistics across pixel, frequency, color, texture, edge, and block dimensions rather than only in distortion strength. VAE is closer on most fidelity measures but shows local edge deviations, whereas SRGAN exhibits larger reconstruction and sharpness deviations, more natural edges, and stronger block artifacts. To diversify the spatial support of these artifacts, we apply mask-aware forgery augmentation to both domains. Given a synthetic image $x_{\mathrm{fake}}$ and its real counterpart $x$, we sample a foreground or background mask $M$ and compute

\begin{equation}
    \hat{x}_{\mathrm{fake}} = M \odot x_{\mathrm{fake}} + (1 - M) \odot x,
\end{equation}
where $\odot$ denotes element-wise multiplication. The resulting image contains localized synthetic regions within real context.

To examine source-specific generalization, we train separate detectors on $T_{vae}$ and $T_{srgan}$ under identical settings. In Fig.~\ref{fig3}, the $T_{vae}$ model (\textbf{\textcolor{blue!80!black}{blue bar}}) performs better on diffusion-based test sets, whereas the $T_{srgan}$ model (\textbf{\textcolor{red!80!black}{red bar}}) is stronger on GAN-based sets. The two aligned sources therefore induce distinct, domain-specific forensic representations.

\subsection{Conflict in Mixed Artifacts Training and Remedy}
\label{theorem}

We next evaluate \textbf{mixed-source training} by aggregating $T_{vae}$ and $T_{srgan}$ into one training set and optimizing a single detector under the same settings. In each batch, real samples come from the shared real-anchor set, while fake samples are randomly mixed from the VAE- and SRGAN-based aligned sources. As shown in Fig.~\ref{fig3} (\textbf{\textcolor{yellow!90!black}{yellow bar}}), direct aggregation does not consistently inherit the complementary strengths of the two single-source models and yields a noticeably weaker overall compromise on several test domains.

This performance degradation stems from the training conflict induced by the significant manifold gap between disparate artifact domains under the shared-real anchor.
Let $\mathcal{D}_{F,V}$ and $\mathcal{D}_{F,S}$ denote the VAE- and SRGAN-based fake subsets of $T_{vae}$ and $T_{srgan}$, respectively; both datasets share the same real subset $\mathcal{D}_R$.
Let $\mathcal{L}(\theta,T)$ denote the average empirical loss of detector parameters $\theta$ on dataset $T$. We define $T_{\mathrm{mix}}$ by sampling from $T_{vae}$ with probability $\lambda\in(0,1)$ and from $T_{srgan}$ with probability $1-\lambda$. Its risk is therefore
\begin{equation}
\mathcal{L}(\theta,T_{\mathrm{mix}})
=\lambda\mathcal{L}(\theta,T_{vae})+(1-\lambda)\mathcal{L}(\theta,T_{srgan}).
\label{eq:mix-risk}
\end{equation}
For $i\in\{vae,srgan\}$, let $T_i$ denote the corresponding aligned source. We define the single-source optimum as $\theta_i^*=\arg\min_{\theta}\mathcal{L}(\theta,T_i)$ and the mixed-source solution as $\theta_{\mathrm{mix}}^*=\arg\min_{\theta}\mathcal{L}(\theta,T_{\mathrm{mix}})$. These ideal optima are used only in the following analysis.

\textbf{\textit{Theorem 1:}} 
Under the shared-real anchor $\mathcal{D}_R$, direct empirical risk minimization (ERM) on $T_{\mathrm{mix}}$ can converge to a sub-optimal mixed-source solution. Evaluating both solutions on the same source $T_i$, we have
\begin{equation}
\mathcal{L}(\theta_{\mathrm{mix}}^*,T_i) > \mathcal{L}(\theta_i^*,T_i).
\label{eq:source-risk-gap}
\end{equation}

\textbf{\textit{Proof 1:}}
For a representative sub-task $T_{vae}$, let $\theta_t$ denote the detector parameters at iteration $t$ and $\eta$ the learning rate, with $\mathbf{g}_{v}=\nabla_{\theta}\mathcal{L}(\theta_t,T_{vae})$ and $\mathbf{g}_{s}=\nabla_{\theta}\mathcal{L}(\theta_t,T_{srgan})$. A mixed-source gradient step gives $\theta_{t+1}=\theta_t-\eta\bigl(\lambda\mathbf{g}_{v}+(1-\lambda)\mathbf{g}_{s}\bigr)$. Applying a first-order Taylor expansion to the VAE risk yields
\begin{multline}
\mathcal{L}(\theta_{t+1},T_{vae}) \approx \mathcal{L}(\theta_t,T_{vae}) - \eta \lambda \|\mathbf{g}_{v}\|^2 - \eta (1-\lambda) \langle \mathbf{g}_{v}, \mathbf{g}_{s} \rangle.
\end{multline}

When the two artifact domains induce incompatible gradients, $\langle \mathbf{g}_{v}, \mathbf{g}_{s} \rangle < 0$ turns the cross-term into an \textbf{adversarial penalty} that raises the source-wise risk. Thus, an aggregate mixed-source update can harm one source even while descending the combined objective.

\begin{figure}[t]
    \centering
    \includegraphics[width=0.95\columnwidth]{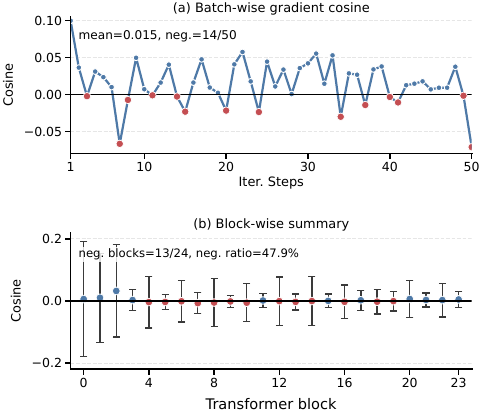}
    \caption{
    \textbf{Gradient conflict between mixed aligned artifacts.} Under direct mixed-source training, VAE- and SRGAN-induced objectives produce unstable gradient alignment. (a) Batch-wise gradient cosine similarity, with red markers denoting negative updates. (b) Block-wise mean and standard deviation across Transformer layers.}
    \label{fig:gradient-conflict}
\end{figure}

We further verify this conflict in Fig.~\ref{fig:gradient-conflict} under the same setting. Across 50 batches of eight samples, we independently compute the VAE- and SRGAN-induced gradients and measure $\cos(\mathbf{g}_{v}, \mathbf{g}_{s})$ over the trainable parameters. The cosine values fluctuate around zero, with a mean of 0.015 and 14 negative steps, showing unstable cross-source cooperation between the two objectives. At the block level, 13 of 24 Transformer blocks have negative mean cosine, with an average negative ratio of 47.9\%. The larger shallow-block variance is consistent with their greater sensitivity to VAE-induced frequency attenuation and SRGAN-induced upsampling patterns. Together, these findings motivate independently training artifact-specific experts and freezing them during cross-artifact fusion \cite{fu2024iisan,fang2025forensic}.

\begin{figure*}[t]
    \centering
    \includegraphics[width=0.95\textwidth]{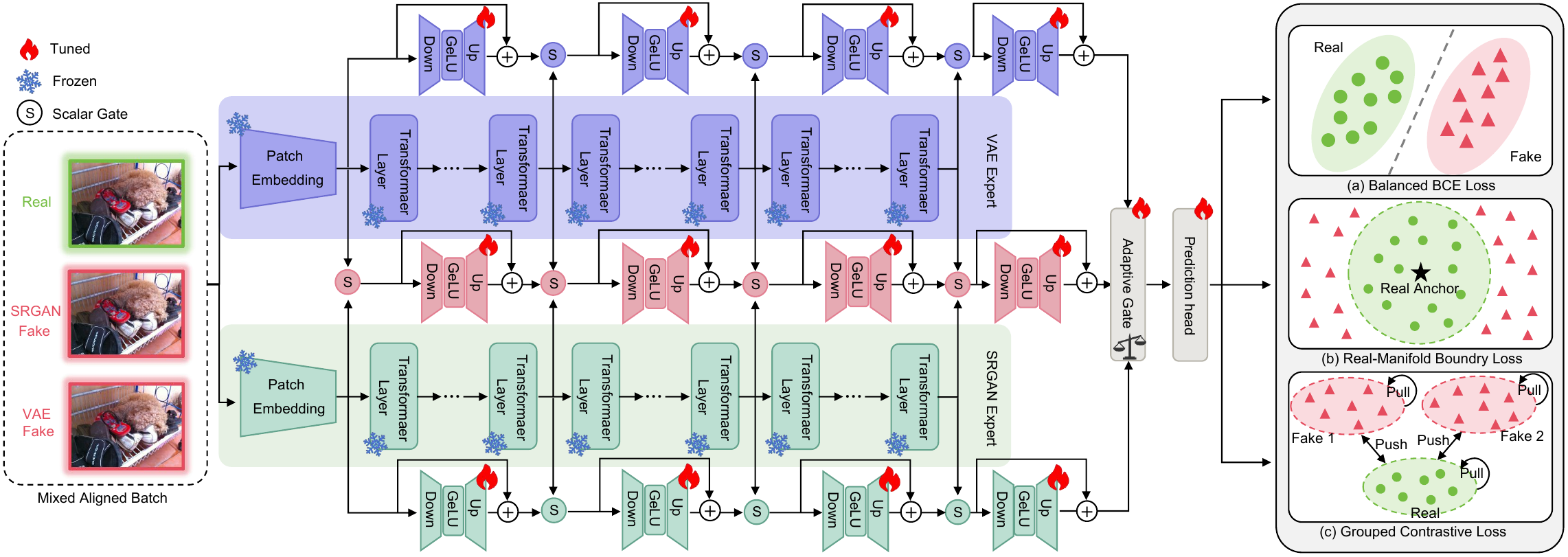}
    \caption{
    \textbf{Layer-wise Fusion of Complementary Experts in ACEF.} In the second stage of ACEF, a mixed aligned batch is processed by frozen VAE and SRGAN experts. LACF reads hidden states from tapped Transformer layers, updates VAE-oriented, SRGAN-oriented, and cross-artifact auxiliary branches, and uses an adaptive branch gate to form the final detection feature.}
    \label{fig4}
\end{figure*} 

\begin{figure}[t]
    \centering
    \includegraphics[width=0.95\columnwidth]{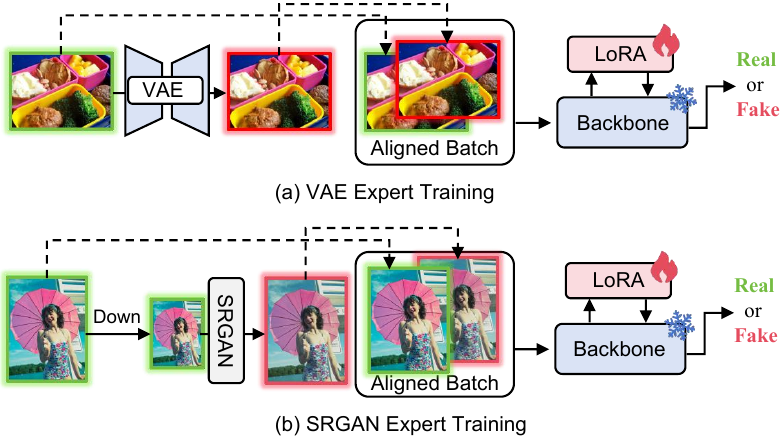}
    \caption{\textbf{Artifact-Specific Expert Training in ACEF.} In the first stage, the VAE expert learns from reconstruction-based aligned pairs, while the SRGAN expert learns from adversarial restoration-based aligned pairs. Both experts adapt a frozen foundation backbone with LoRA, enabling each branch to focus on one aligned artifact source before fusion.}
    \label{fig:expert}
\end{figure}
\textbf{\textit{Theorem 2:}}
Let $F_{\mathrm{dec}}(x; \phi) = \pi_v(x; \phi)E_{vae}(x; \theta_{vae}^*) + \pi_s(x; \phi)E_{srgan}(x; \theta_{srgan}^*)$ denote fusion with frozen source-specific experts, where $\phi$ contains the trainable fusion parameters. If the frozen-expert representations for $\mathcal{D}_{F,V}$ and $\mathcal{D}_{F,S}$ are separable, there exists an optimal $\phi^*$ such that the decoupled risk is lower than direct mixed-source ERM
\begin{equation}
\mathcal{L}(\phi^*,T_{\mathrm{mix}}) < \mathcal{L}(\theta_{\mathrm{mix}}^*,T_{\mathrm{mix}}),
\end{equation}
where $\theta_{\mathrm{mix}}^*$ is the mixed-source solution from Theorem 1.

\textbf{\textit{Proof 2:}}
The empirical risk over $T_{\mathrm{mix}}$, expressed as an expectation over uniformly sampled examples, is
\begin{equation}
\begin{split}
\mathcal{L}(\phi,T_{\mathrm{mix}})
&= \mathbb{E}_{(x,y) \sim T_{\mathrm{mix}}} \bigg[
\ell \bigg( \pi_v(x; \phi)E_{vae}(x; \theta_{vae}^*) \\
&\quad + \pi_s(x; \phi)E_{srgan}(x; \theta_{srgan}^*),\, y \bigg)
\bigg].
\end{split}
\end{equation}
Freezing experts at $\{\theta_i^*\}$ restricts optimization to $\phi$ and isolates both experts from the antagonistic updates identified in Theorem 1.

For theoretical analysis, we assume an oracle hard gate that routes VAE- and SRGAN-aligned samples to their corresponding experts. Shared real anchors follow the source component of their aligned pairs. With routing error excluded, the total risk admits the following decomposition:
\begin{equation}
\mathcal{L}(\phi^*,T_{\mathrm{mix}}) = \lambda \mathcal{L}(\theta_{vae}^*,T_{vae}) + (1-\lambda) \mathcal{L}(\theta_{srgan}^*,T_{srgan}).
\label{eq:dec-risk}
\end{equation}

Combining Eqs.~\eqref{eq:source-risk-gap}, \eqref{eq:dec-risk}, and \eqref{eq:mix-risk} gives $\mathcal{L}(\phi^*,T_{\mathrm{mix}}) < \mathcal{L}(\theta_{\mathrm{mix}}^*,T_{\mathrm{mix}})$, establishing the claimed advantage over direct mixed-source ERM.

This idealized result motivates decoupled learning, but hard source routing is impractical because artifact sources are unknown and may overlap at inference. We therefore introduce a learnable realization, \textbf{Artifact-Complementary Expert Fusion (ACEF)}. As shown in Fig.~\ref{fig3} (\textbf{\textcolor{green!80!black}{green bar}}), ACEF outperforms direct mixed-source training and also surpasses the SRGAN expert on GAN-based generators and the VAE expert on diffusion-based generators. This indicates that complementary fusion improves both domains without sacrificing expert specialization.

\subsection{Artifact-Complementary Expert Fusion}
ACEF follows a two-stage design that separates artifact specialization from cross-artifact integration. First, two experts are independently adapted to VAE reconstruction and SRGAN restoration artifacts. In the second stage, both experts are frozen, and LACF learns to combine their multi-layer source-specific and cross-artifact evidence through lightweight gates.

\subsubsection{Artifact-Specific Expert Training}
As shown in Fig.~\ref{fig:expert}, we independently train $E_{vae}$ on $T_{vae}$ and $E_{srgan}$ on $T_{srgan}$. For each domain $i\in\{vae,srgan\}$, the expert comprises a frozen backbone $\Phi$, a domain-specific LoRA adapter \cite{hu2021lora} with parameters $\Delta\theta_i$, and a binary prediction head; only the adapter and head are optimized. For a labeled sample $(x_n,y_n)$, where $y_n=0$ denotes real and $y_n=1$ fake, its expert feature is $\mathbf{f}_n^i=\Phi(x_n;\Delta\theta_i)$. The prediction head is supervised by a standard binary cross-entropy loss $\mathcal{L}_{\mathrm{BCE}}^i$.

To structure the learned feature space, we also apply a margin-based contrastive loss. We normalize each feature as $\bar{\mathbf{f}}_n^i=\mathbf{f}_n^i/\|\mathbf{f}_n^i\|_2$ and define $d_{pq}^i=\|\bar{\mathbf{f}}_p^i-\bar{\mathbf{f}}_q^i\|_2$. The sets $\mathcal{P}_i^+$ and $\mathcal{P}_i^-$ contain pairs with identical and different labels, respectively. The loss is
\begin{equation}
\label{eq:expert-con}
\mathcal{L}_{\mathrm{con}}^i=
\mathbb{E}_{(p,q)\in\mathcal{P}_i^+}[d_{pq}^i-m^+]_+
+
\mathbb{E}_{(p,q)\in\mathcal{P}_i^-}[m^- - d_{pq}^i]_+,
\end{equation}
where $[z]_+=\max(z,0)$, $m^+=0$, and $m^-=1$. The expert objective is
\begin{equation}
\mathcal{L}_i=\lambda_{\mathrm{BCE}}\mathcal{L}_{\mathrm{BCE}}^i+\lambda_{\mathrm{con}}\mathcal{L}_{\mathrm{con}}^i,
\end{equation}
with $\lambda_{\mathrm{BCE}}=\lambda_{\mathrm{con}}=0.5$.

\subsubsection{Layer-wise Expert Fusion}
As illustrated in Fig.~\ref{fig4}, \textbf{Layer-wise Artifact-Complementary Fusion (LACF)} reads intermediate hidden states from the frozen experts because final global descriptors may miss depth-distributed forensic evidence. It trains only lightweight auxiliary branches, scalar layer gates, an adaptive branch gate, and the prediction head.

Let $\mathcal{K}=\{k_1,\ldots,k_{M_{\mathrm{tap}}}\}$ denote the tapped Transformer layers, selected at equal intervals along the network depth. For an input image $x$, the frozen experts output hidden states $\mathbf{h}_{vae}^{m}$ and $\mathbf{h}_{srgan}^{m}$ at layer $k_m$. Let $\mathbf{z}_{q}^{m}$ denote the accumulated fusion state of branch $q\in\{vae,srgan,c\}$ after processing the $m$-th tapped layer. We initialize $\mathbf{z}_{vae}^{0}$, $\mathbf{z}_{srgan}^{0}$, and $\mathbf{z}_{c}^{0}$ as zero vectors. Each scalar gate is parameterized as $r_q^m=\sigma(a_q^m/\tau)$ for $q\in\{vae,srgan,c\}$, where $\sigma(\cdot)$ denotes the sigmoid function, $a_q^m$ is a learnable logit, and $\tau$ is the temperature. LACF first updates two source-oriented branch states as
\begin{equation}
\mathbf{z}_{j}^{m}=A_{j}^{m}\bigl(r_{j}^{m}\mathbf{h}_{j}^{m}+(1-r_{j}^{m})\mathbf{z}_{j}^{m-1}\bigr),\quad j\in\{vae,srgan\}.
\end{equation}
Here $A_j^m(\cdot)$ is a bottleneck fusion block composed of down projection, nonlinearity (e.g., GeLU), up projection, and a residual connection. The scalar gate $r_j^m\in[0,1]$ balances the current hidden state and the accumulated branch state from shallower tapped layers. In parallel, LACF maintains a cross-artifact branch:
\begin{equation}
\mathbf{z}_{c}^{m}=A_{c}^{m}\bigl(\mathbf{z}_{c}^{m-1}+r_{c}^{m}\mathbf{h}_{vae}^{m}+(1-r_{c}^{m})\mathbf{h}_{srgan}^{m}\bigr).
\end{equation}
The scalar gate $r_c^m\in[0,1]$ controls contributions of VAE-oriented and SRGAN-oriented evidence before updating the shared cross-artifact state. Thus, source-oriented branches retain expert-specific traces, while the cross branch models their interaction at matching semantic depths.

After the last tapped layer, the branch outputs are denoted as $\tilde{\mathbf{z}}_{vae}$, $\tilde{\mathbf{z}}_{srgan}$, and $\tilde{\mathbf{z}}_{c}$. A sample-adaptive branch gate predicts their fusion weights:
\begin{equation}
\boldsymbol{\omega}=\mathrm{softmax}\bigl(G_b(\operatorname{Concat}(\tilde{\mathbf{z}}_{vae},\tilde{\mathbf{z}}_{srgan},\tilde{\mathbf{z}}_{c}))\bigr).
\end{equation}
Here $G_b(\cdot)$ is the Adaptive Gate and $\boldsymbol{\omega}=(\omega_{vae},\omega_{srgan},\omega_c)$ denotes sample-specific weights. The fused feature is computed by
\begin{equation}
\mathbf{u}=\omega_{vae}\tilde{\mathbf{z}}_{vae}+\omega_{srgan}\tilde{\mathbf{z}}_{srgan}+\omega_{c}\tilde{\mathbf{z}}_{c}.
\end{equation}
The binary prediction head takes $\mathbf{u}$ as input. During fusion training, each mini-batch contains real images, VAE-aligned fakes, and SRGAN-aligned fakes. We denote this triplet batch as $\mathcal{B}_{\mathrm{tri}}=\{(x_n,y_n,s_n)\}_{n=1}^{N_{\mathrm{fus}}}$, where $s_n\in\{\mathrm{real},\mathrm{vae},\mathrm{srgan}\}$ is the source label and $\mathbf{u}_n$ is the fused feature of $x_n$. We use $N_-$ and $N_+$ to denote the numbers of real ($y_n=0$) and fake ($y_n=1$) samples in the current batch, respectively.

The prediction head maps $\mathbf{u}_n$ to logit $b_n$. We use a balanced BCE loss $\mathcal{L}_{\mathrm{cls}}$ with positive weight $w^+=N_-/N_+$.
For each fake source $s\in\mathcal{S}=\{\mathrm{vae},\mathrm{srgan}\}$, we form $\mathcal{G}_s=\{n\mid y_n=0 \text{ or }s_n=s\}$ from all real samples and only source-$s$ fakes. The grouped contrastive loss is
\begin{equation}
\mathcal{L}_{\mathrm{gcon}}=\frac{1}{|\mathcal{S}|}\sum_{s\in\mathcal{S}}\mathcal{C}(\{(\mathbf{u}_n,y_n)\}_{n\in\mathcal{G}_s}),
\end{equation}
where $\mathcal{C}(\cdot)$ is the contrastive loss in Eq.~\eqref{eq:expert-con}. This source-wise grouping contrasts each fake domain against the shared real 
samples without forcing heterogeneous fakes into one cluster.

We further add a real-manifold boundary loss because aligned fakes may remain close to the real cluster after binary supervision. We normalize each feature as $\bar{\mathbf{u}}_n=\mathbf{u}_n/\|\mathbf{u}_n\|_2$ and define $\mathbf{c}_r$ as the normalized mean of real features. The boundary loss is
\begin{equation}
\mathcal{L}_{\mathrm{bd}}=\frac{1}{N_-}\sum_{n:y_n=0}(1-\bar{\mathbf{u}}_n^\top\mathbf{c}_r)+\frac{1}{N_+}\sum_{n:y_n=1}[\bar{\mathbf{u}}_n^\top\mathbf{c}_r-\rho]_+^2.
\end{equation}
This term keeps real samples compact around the real manifold and penalizes fake samples whose cosine similarity to the real center exceeds the margin $\rho$. We set $\rho=0.3$ in our experiments.
The final fusion objective is
\begin{equation}
\mathcal{L}_{\mathrm{fuse}}=\lambda_{\mathrm{cls}}\mathcal{L}_{\mathrm{cls}}+\lambda_{\mathrm{g}}\mathcal{L}_{\mathrm{gcon}}+\lambda_{\mathrm{b}}\mathcal{L}_{\mathrm{bd}}.
\end{equation}
We use $\lambda_{\mathrm{cls}}=0.5$, $\lambda_{\mathrm{g}}=0.25$, and $\lambda_{\mathrm{b}}=0.25$ by default.

\section{Experiments}

\begin{table*}[tb!]
\centering
\caption{\textbf{Cross-domain forgery benchmark with GAN- and diffusion-based generators.} GAN Avg., Diff. Avg., and Avg. Acc. expose artifact-family balance across generator groups. Values are balanced accuracy (\%) across all subsets. The best results are in \textbf{bold}, and the second-best are \underline{underlined}. 
}
\label{tab:cross-forgery}
\begingroup
\scriptsize
\setlength{\tabcolsep}{1.2pt}
\renewcommand{\arraystretch}{0.84}
\begin{tabularx}{\linewidth}{
    >{\raggedright\arraybackslash}p{1.55cm}
    *{20}{C}
    @{\hspace{1.8pt}}
    !{\color{black!30}\vrule width 0.45pt}
    @{\hspace{1.8pt}}
    C
}
\toprule
\multirow{2}{*}{\textbf{Method}}
& \multicolumn{10}{c}{\textbf{GAN-based generators}}
& \multicolumn{10}{c}{\textbf{Diffusion-based generators}}
& \multirow{2}{*}{\makecell{\textbf{Avg.}\\\textbf{Acc.}}}
\\
\cmidrule(lr){2-11} \cmidrule(lr){12-21}
& \makecell{Big\\GAN}
& \makecell{Cycle\\GAN}
& \makecell{Pro\\GAN}
& \makecell{Star\\GAN}
& \makecell{GAN\\GAN}
& \makecell{R3\\GAN}
& \makecell{SG-\\XL}
& SG3
& \makecell{Style\\Swim}
& \makecell{\textbf{GAN}\\\textbf{Avg.}}
& VQDM
& ADM
& GLIDE
& \makecell{SD-\\XL}
& \makecell{MJ-\\V6}
& SD3
& \makecell{DALL\\-E3}
& FLUX
& \makecell{Ima\\gen3}
& \makecell{\textbf{Diff.}\\\textbf{Avg.}}
&
\\
\midrule
NPR  &  $84.3$  &  $96.1$  &  $99.8$  &  $99.3$  &  $82.5$  &  $77.7$  &  $75.8$  &  $78.9$  &  $77.0$  &  $85.7$  &  $78.1$  &  $69.7$  &  $79.8$  &  $78.3$  &  $61.3$  &  $78.3$  &  $66.6$  &  $79.5$  &  $79.5$  &  $74.6$  &  $80.2$\\
UnivFD  &  $95.1$  &  $98.3$  &  $99.8$  &  $95.8$  &  $\underline{99.5}$  &  $78.5$  &  $80.5$  &  $61.1$  &  $84.1$  &  $88.1$  &  $85.4$  &  $97.7$  &  $59.9$  &  $59.1$  &  $52.3$  &  $56.3$  &  $53.5$  &  $48.4$  &  $50.6$  &  $62.6$  &  $75.4$\\
FatFormer  &  $\mathbf{99.5}$  &  $\underline{99.4}$  &  $\underline{99.9}$  &  $\underline{99.8}$  &  $99.4$  &  $94.4$  &  $93.4$  &  $90.7$  &  $93.9$  &  $\mathbf{96.7}$  &  $86.9$  &  $78.5$  &  $84.4$  &  $75.0$  &  $47.6$  &  $62.9$  &  $50.1$  &  $48.2$  &  $46.9$  &  $64.5$  &  $80.6$\\
SAFE  &  $89.7$  &  $98.9$  &  $\underline{99.9}$  &  $\mathbf{99.9}$  &  $91.5$  &  $91.6$  &  $91.0$  &  $90.4$  &  $91.6$  &  $93.8$  &  $96.3$  &  $82.1$  &  $90.5$  &  $93.3$  &  $76.1$  &  $92.4$  &  $41.9$  &  $91.4$  &  $89.6$  &  $83.7$  &  $88.8$\\
C2P-CLIP  &  $\underline{99.0}$  &  $\mathbf{99.7}$  &  $\mathbf{100.0}$  &  $\underline{99.8}$  &  $\mathbf{99.7}$  &  $92.5$  &  $58.1$  &  $94.4$  &  $91.2$  &  $92.7$  &  $79.0$  &  $65.1$  &  $87.1$  &  $74.1$  &  $60.6$  &  $67.7$  &  $50.4$  &  $57.9$  &  $49.5$  &  $65.7$  &  $79.2$\\
AIDE  &  $84.9$  &  $92.0$  &  $92.8$  &  $88.1$  &  $87.7$  &  $90.8$  &  $88.3$  &  $85.6$  &  $82.1$  &  $88.0$  &  $\mathbf{99.9}$  &  $\mathbf{99.6}$  &  $\mathbf{94.3}$  &  $93.3$  &  $74.8$  &  $92.9$  &  $66.3$  &  $89.8$  &  $90.5$  &  $89.0$  &  $88.5$\\
DRCT  &  $82.2$  &  $89.4$  &  $70.4$  &  $53.0$  &  $83.3$  &  $44.0$  &  $49.0$  &  $65.2$  &  $72.2$  &  $67.6$  &  $89.1$  &  $75.2$  &  $69.2$  &  $62.0$  &  $64.7$  &  $57.7$  &  $73.8$  &  $51.4$  &  $53.0$  &  $66.2$  &  $66.9$\\
Aligned  &  $66.5$  &  $55.3$  &  $74.5$  &  $77.6$  &  $68.6$  &  $67.7$  &  $56.4$  &  $48.3$  &  $68.4$  &  $64.8$  &  $71.1$  &  $63.1$  &  $51.3$  &  $86.6$  &  $83.3$  &  $78.8$  &  $81.5$  &  $78.4$  &  $81.6$  &  $75.1$  &  $70.0$\\
\midrule
DDA~(v2-L)  &  $83.6$  &  $63.9$  &  $92.3$  &  $66.3$  &  $89.9$  &  $\mathbf{96.0}$  &  $48.1$  &  $60.1$  &  $52.5$  &  $72.5$  &  $71.8$  &  $88.0$  &  $84.7$  &  $\mathbf{97.8}$  &  $\mathbf{96.6}$  &  $\mathbf{97.7}$  &  $\mathbf{94.9}$  &  $\underline{93.0}$  &  $85.7$  &  $90.0$  &  $81.3$\\
\rowcolor{oursrow} \textbf{Ours}~(v2-L)  &  $96.0$  &  $94.5$  &  $85.9$  &  $75.5$  &  $97.7$  &  $\underline{94.9}$  &  $78.6$  &  $92.5$  &  $92.6$  &  $89.8$  &  $91.9$  &  $94.7$  &  $92.5$  &  $95.3$  &  $95.3$  &  $95.3$  &  $93.6$  &  $85.0$  &  $86.2$  &  $92.2$  &  $91.0$\\
\addlinespace[1.5pt]
DDA~(v3-L)  &  $95.7$  &  $91.6$  &  $92.0$  &  $73.9$  &  $96.5$  &  $90.5$  &  $82.3$  &  $91.7$  &  $89.1$  &  $89.2$  &  $97.8$  &  $95.5$  &  $88.3$  &  $91.5$  &  $95.4$  &  $91.5$  &  $91.1$  &  $91.0$  &  $90.6$  &  $92.5$  &  $90.9$\\
\rowcolor{oursrow} \textbf{Ours}~(v3-L)  &  $97.0$  &  $96.0$  &  $92.0$  &  $82.5$  &  $98.0$  &  $\underline{94.9}$  &  $\mathbf{94.7}$  &  $\underline{96.9}$  &  $\mathbf{94.8}$  &  $94.1$  &  $98.4$  &  $96.8$  &  $\underline{94.2}$  &  $94.9$  &  $94.8$  &  $94.9$  &  $94.3$  &  $\mathbf{94.3}$  &  $\underline{93.7}$  &  $\underline{95.1}$  &  $\underline{94.6}$\\
\addlinespace[1.5pt]
DDA~(v3-H+)  &  $97.7$  &  $95.6$  &  $90.0$  &  $94.4$  &  $97.5$  &  $89.3$  &  $89.0$  &  $84.8$  &  $86.2$  &  $91.6$  &  $\underline{99.0}$  &  $98.1$  &  $88.6$  &  $91.4$  &  $90.2$  &  $91.4$  &  $89.3$  &  $88.7$  &  $88.7$  &  $91.7$  &  $91.7$\\
\rowcolor{oursrow} \textbf{Ours}~(v3-H+)  &  $97.8$  &  $97.7$  &  $91.2$  &  $89.1$  &  $98.9$  &  $94.8$  &  $\underline{94.4}$  &  $\mathbf{98.0}$  &  $\underline{94.4}$  &  $\underline{95.1}$  &  $98.9$  &  $\underline{98.7}$  &  $\mathbf{94.3}$  &  $\underline{95.4}$  &  $\underline{96.2}$  &  $\underline{95.4}$  &  $\underline{94.4}$  &  $\mathbf{94.3}$  &  $\mathbf{94.4}$  &  $\mathbf{95.8}$  &  $\mathbf{95.5}$\\
\bottomrule
\end{tabularx}
\endgroup
\end{table*}

\subsection{Implementation Details}
Our framework uses DINO visual backbones \cite{caron2021emerging,oquab2024dinov2,simeoni2025dinov3} adapted with LoRA \cite{hu2021lora}, with rank $r=8$ and scaling factor $\alpha=1$. We report results with DINO-v2-L, DINO-v3-L, and DINO-v3-H+. Following DDA \cite{chen2025dual}, training images are resized to $336 \times 336$ and randomly cropped, whereas evaluation uses center cropping. Training pairs are built from MSCOCO real images \cite{lin2014microsoft} and two aligned negatives generated by VAE reconstruction and SRGAN upsampling. To preserve pixel-level correspondence between real and synthetic pairs, images are padded or cropped to multiples of 8 before synthesis, and SRGAN negatives use $4\times$ bicubic downsampling as $D(\cdot)$. Training has two stages. We train the VAE and SRGAN experts separately for 10,000 iterations with batch size 16 and four-step gradient accumulation. We then freeze both experts and train only LACF for 5,000 iterations with batch size 32 and the same gradient accumulation. Unless otherwise specified, LACF taps six Transformer layers with bottleneck width $d_b=128$, dropout 0.1, and temperature $\tau=0.2$. We use a learning rate of $1 \times 10^{-4}$ with cosine annealing and train on 8 NVIDIA V100 GPUs. The threshold is fixed at 0.5.

\subsection{Benchmarks and Comparative Methods}
We evaluate generalization at three levels. First, we use a cross-domain forgery benchmark reorganized from \cite{li2025artificial,zhu2023genimage,wang2020cnn}, grouping test generators into GAN- and diffusion-based families to assess artifact-family generalization. Second, we evaluate cross-benchmark transfer on GenImage \cite{zhu2023genimage}, DRCT-2M \cite{chen2024drct}, Synthbuster \cite{bammey2023synthbuster}, DDA-COCO \cite{chen2025dual}, EvalGEN \cite{chen2025dual}, AIGCDetectionBenchmark \cite{zhong2023patchcraft}, ForenSynths \cite{wang2020cnn}, Chameleon \cite{yan2024sanity}, WildRF \cite{cavia2024real}, SynthWildx \cite{cozzolino2024raising}, and BFree-Online \cite{guillaro2025bias}, which cover generator-centric, online-source, recent diffusion, and in-the-wild distributions. Third, we evaluate localized forgery on the inpainting benchmark BR-GEN \cite{cai2026zooming}. We randomly sample 3,000 fake images from each subcategory with their corresponding real images and inspect every sample to ensure that the forged region remains visible after center cropping. We compare low-level methods NPR \cite{tan2024rethinking}, SAFE \cite{li2025improving}, and AIDE \cite{yan2024sanity}; foundation-model-based methods UnivFD \cite{ojha2023towards}, FatFormer \cite{liu2024forgery}, and C2P-CLIP \cite{tan2025c2p}; and aligned methods DRCT \cite{chen2024drct}, Aligned \cite{rajan2024aligned}, and DDA \cite{chen2025dual}. We use official checkpoints when available and otherwise report published results. Following DDA \cite{chen2025dual}, synthetic images in GenImage, ForenSynths, and AIGCDetectionBenchmark are JPEG-compressed with quality factor 96.

\begin{figure*}[!t]
    \centering
    \includegraphics[width=0.95\textwidth]{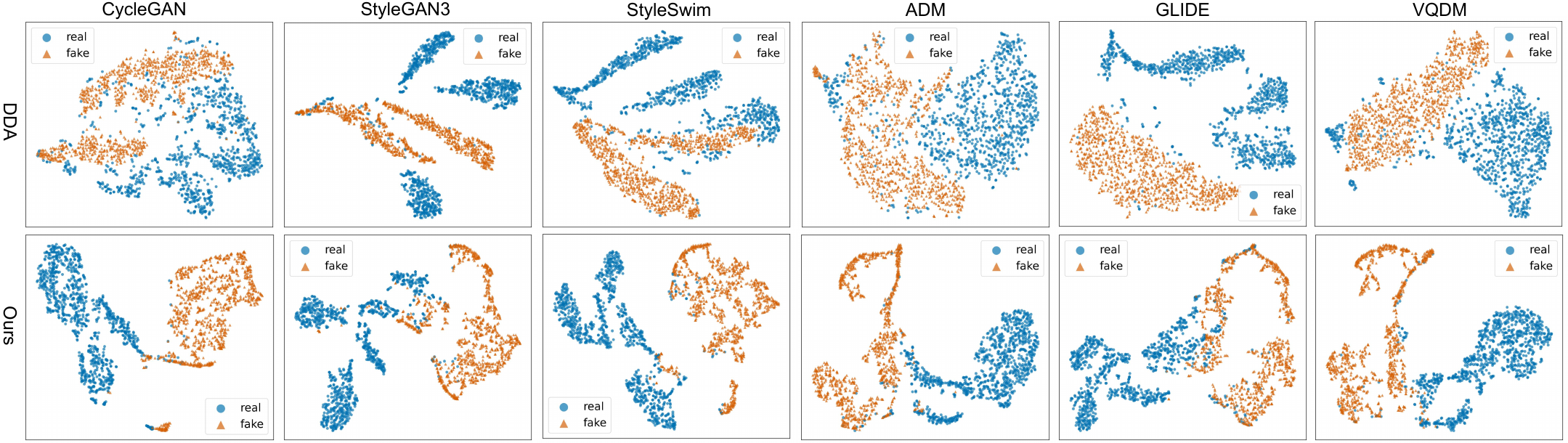}
    \caption{\textbf{Feature-space comparison between DDA and Ours with DINO-v2-L.} Blue circles and orange triangles denote real and fake samples, respectively. Columns cover three GAN-based and three diffusion-based generators.}
    \label{fig:tsne}
\end{figure*}

\subsection{Performance}

\begin{table*}[t]
\centering
\caption{
\textbf{Cross-benchmark generalization across diverse evaluation datasets.}
}
\label{tab:compare-methods}
\begingroup
\scriptsize
\setlength{\tabcolsep}{1.6pt}
\renewcommand{\arraystretch}{0.84}
\begin{tabularx}{\linewidth}{
    >{\raggedright\arraybackslash}p{1.78cm}
    *{11}{C}
    @{\hspace{1.8pt}}
    !{\color{black!30}\vrule width 0.45pt}
    @{\hspace{1.8pt}}
    C
}
\toprule
\textbf{Method}
& \makecell{GenImage}
& \makecell{DRCT-\\2M}
& \makecell{DDA-\\COCO}
& \makecell{EvalGEN}
& \makecell{Synth\\buster}
& \makecell{Foren\\Synths}
& \makecell{AIGC\\Detect.}
& \makecell{Chameleon}
& \makecell{Synth\\Wildx}
& \makecell{WildRF}
& \makecell{BF-\\Online}
& \textbf{Avg.}
\\
\midrule
NPR
& $51.5$ & $37.3$ & $42.2$ & $2.9$
& $50.0$ & $47.9$ & $53.1$ & $59.9$
& $49.8$ & $63.5$ & $49.5$ & $46.1$
\\
UnivFD
& $64.1$ & $61.8$ & $52.4$ & $15.4$
& $67.8$ & $77.7$ & $72.5$ & $50.7$
& $52.3$ & $55.3$ & $49.0$ & $56.3$
\\
FatFormer
& $62.8$ & $52.2$ & $51.7$ & $45.6$
& $56.1$ & $90.1$ & $85.0$ & $51.2$
& $52.1$ & $58.9$ & $50.0$ & $59.6$
\\
SAFE
& $50.3$ & $59.3$ & $49.9$ & $1.1$
& $46.5$ & $49.7$ & $50.3$ & $59.2$
& $49.1$ & $54.2$ & $50.5$ & $47.3$
\\
C2P-CLIP
& $74.4$ & $59.2$ & $51.3$ & $38.9$
& $68.5$ & $\underline{92.1}$ & $81.4$ & $51.1$
& $57.1$ & $59.6$ & $50.0$ & $62.1$
\\
AIDE
& $61.2$ & $64.6$ & $50.0$ & $19.1$
& $53.9$ & $59.4$ & $63.6$ & $63.1$
& $54.7$ & $58.4$ & $53.1$ & $54.6$
\\
DRCT
& $85.3$ & $90.5$ & $60.2$ & $77.8$
& $81.3$ & $73.9$ & $81.4$ & $56.6$
& $55.1$ & $51.6$ & $55.7$ & $69.9$
\\
Aligned
& $79.0$ & $95.5$ & $86.5$ & $68.0$
& $77.4$ & $53.9$ & $66.6$ & $71.0$
& $78.8$ & $80.1$ & $68.5$ & $75.0$
\\
\midrule
DDA~(v2-L)
& $91.7$ & $98.1$ & $92.2$ & $97.2$
& $90.1$ & $81.4$ & $87.8$ & $82.4$
& $90.9$ & $90.3$ & $\mathbf{95.1}$ & $90.7$
\\

\rowcolor{oursrow}
\textbf{Ours}~(v2-L)
& $96.5$ & $98.4$ & $95.0$ & $93.8$
& $\mathbf{97.6}$ & $90.9$ & $93.8$ & $82.6$
& $\underline{91.9}$ & $92.8$ & $93.2$ & $93.3$
\\

\addlinespace[1.5pt]

DDA~(v3-L)
& $97.7$ & $98.9$ & $96.2$ & $\mathbf{99.5}$
& $90.6$ & $85.4$ & $94.9$ & $90.8$
& $87.4$ & $91.7$ & $90.5$ & $93.1$
\\

\rowcolor{oursrow}
\textbf{Ours}~(v3-L)
& $\underline{98.2}$ & $99.0$ & $96.1$ & $97.8$
& $92.9$ & $91.6$ & $96.2$ & $87.9$
& $91.3$ & $\underline{95.6}$ & $90.3$ & $94.3$
\\

\addlinespace[1.5pt]

DDA~(v3-H+)
& $\mathbf{98.8}$ & $\underline{99.3}$ & $\underline{96.7}$ & $\underline{99.0}$
& $\underline{96.8}$ & $88.5$ & $\underline{96.5}$ & $\underline{93.3}$
& $88.3$ & $93.6$ & $91.1$ & $\underline{94.7}$
\\

\rowcolor{oursrow}
\textbf{Ours}~(v3-H+)
& $\mathbf{98.8}$ & $\mathbf{99.4}$ & $\mathbf{97.5}$ & $98.8$
& $95.8$ & $\mathbf{93.9}$ & $\mathbf{97.1}$ & $\mathbf{93.5}$
& $\mathbf{92.2}$ & $\mathbf{95.9}$ & $\underline{94.6}$ & $\mathbf{96.1}$
\\

\bottomrule
\end{tabularx}

\endgroup
\end{table*}

\begin{table*}[tb!]
  \centering
  \caption{\textbf{Performance Comparison on the DRCT-2M Benchmarks.}}
  \label{tab:compare-drct}
\begingroup
\scriptsize
\setlength{\tabcolsep}{1.0pt}
\renewcommand{\arraystretch}{0.84}
\begin{tabularx}{\linewidth}{
    >{\raggedright\arraybackslash}p{1.55cm}
    *{16}{C}
    @{\hspace{1.8pt}}
    !{\color{black!30}\vrule width 0.45pt}
    @{\hspace{1.8pt}}
    C
}
\toprule
\textbf{Method}
& LDM
& \makecell{SD\\1.4}
& \makecell{SD\\1.5}
& SD2
& SDXL
& \makecell{SDXL\\Ref.}
& \makecell{SD\\Turbo}
& \makecell{SDXL\\Turbo}
& \makecell{LCM\\SD1.5}
& \makecell{LCM\\SDXL}
& \makecell{SD1\\Ctrl}
& \makecell{SD2\\Ctrl}
& \makecell{SDXL\\Ctrl}
& \makecell{SD1\\DR}
& \makecell{SD2\\DR}
& \makecell{SDXL\\DR}
& \textbf{Avg.}
\\
\midrule
NPR  &  $33.0$  &  $29.1$  &  $29.0$  &  $35.1$  &  $33.2$  &  $28.4$  &  $27.9$  &  $27.9$  &  $29.4$  &  $30.2$  &  $28.4$  &  $28.3$  &  $34.7$  &  $67.9$  &  $67.4$  &  $66.1$  &  $37.3$\\
UnivFD  &  $85.4$  &  $56.8$  &  $56.4$  &  $58.2$  &  $63.2$  &  $55.0$  &  $56.5$  &  $53.0$  &  $54.5$  &  $65.9$  &  $68.0$  &  $65.4$  &  $75.9$  &  $64.6$  &  $56.2$  &  $53.9$  &  $61.8$\\
FatFormer  &  $55.9$  &  $48.2$  &  $48.2$  &  $48.2$  &  $48.2$  &  $48.3$  &  $48.2$  &  $48.2$  &  $48.3$  &  $50.6$  &  $49.7$  &  $49.9$  &  $59.8$  &  $66.3$  &  $60.6$  &  $56.0$  &  $52.2$\\
SAFE  &  $50.3$  &  $50.1$  &  $50.0$  &  $50.0$  &  $49.9$  &  $50.1$  &  $50.0$  &  $50.0$  &  $50.1$  &  $50.0$  &  $49.9$  &  $50.0$  &  $54.7$  &  $98.2$  &  $98.5$  &  $97.3$  &  $59.3$\\
C2P-CLIP  &  $83.0$  &  $51.7$  &  $51.7$  &  $52.9$  &  $51.9$  &  $64.6$  &  $51.7$  &  $50.6$  &  $52.0$  &  $66.1$  &  $56.9$  &  $54.7$  &  $77.8$  &  $67.2$  &  $57.1$  &  $56.7$  &  $59.2$\\
AIDE  &  $64.4$  &  $74.9$  &  $75.1$  &  $58.5$  &  $53.5$  &  $66.3$  &  $52.8$  &  $52.8$  &  $70.0$  &  $54.3$  &  $65.9$  &  $53.6$  &  $53.9$  &  $95.3$  &  $73.3$  &  $69.0$  &  $64.6$\\
DRCT  &  $96.7$  &  $96.3$  &  $96.3$  &  $94.9$  &  $96.2$  &  $93.5$  &  $93.4$  &  $92.9$  &  $91.2$  &  $95.0$  &  $95.6$  &  $92.7$  &  $92.0$  &  $94.1$  &  $69.6$  &  $57.4$  &  $90.5$\\
Aligned  &  $\mathbf{99.9}$  &  $\mathbf{99.9}$  &  $\mathbf{99.9}$  &  $\mathbf{99.6}$  &  $90.2$  &  $81.3$  &  $\mathbf{99.7}$  &  $89.4$  &  $\mathbf{99.7}$  &  $90.0$  &  $\mathbf{99.9}$  &  $\underline{99.2}$  &  $87.6$  &  $\mathbf{99.9}$  &  $\mathbf{99.8}$  &  $92.6$  &  $95.5$\\
\midrule
DDA~(v2-L)  &  $99.2$  &  $98.9$  &  $99.0$  &  $98.3$  &  $98.0$  &  $96.8$  &  $97.9$  &  $94.8$  &  $95.9$  &  $98.2$  &  $98.7$  &  $99.0$  &  $\mathbf{99.4}$  &  $99.0$  &  $\underline{99.5}$  &  $96.3$  &  $98.1$\\
\rowcolor{oursrow} \textbf{Ours}~(v2-L)  &  $98.7$  &  $98.4$  &  $98.4$  &  $98.4$  &  $98.2$  &  $97.9$  &  $98.4$  &  $97.6$  &  $97.9$  &  $98.7$  &  $98.6$  &  $98.6$  &  $98.7$  &  $98.7$  &  $98.7$  &  $98.2$  &  $98.4$\\
\addlinespace[1.5pt]
DDA~(v3-L)  &  $99.3$  &  $98.9$  &  $98.8$  &  $98.7$  &  $\underline{99.1}$  &  $98.5$  &  $98.7$  &  $98.4$  &  $98.9$  &  $\underline{99.2}$  &  $99.3$  &  $\underline{99.2}$  &  $\underline{99.3}$  &  $99.3$  &  $99.2$  &  $98.2$  &  $98.9$\\
\rowcolor{oursrow} \textbf{Ours}~(v3-L)  &  $99.3$  &  $98.9$  &  $98.8$  &  $98.6$  &  $99.0$  &  $98.1$  &  $98.9$  &  $\underline{98.7}$  &  $98.9$  &  $\underline{99.2}$  &  $99.3$  &  $\underline{99.2}$  &  $\underline{99.3}$  &  $99.3$  &  $99.3$  &  $98.9$  &  $99.0$\\
\addlinespace[1.5pt]
DDA~(v3-H+)  &  $\underline{99.4}$  &  $99.2$  &  $99.1$  &  $\underline{99.4}$  &  $\mathbf{99.2}$  &  $\underline{99.1}$  &  $\underline{99.4}$  &  $\mathbf{99.4}$  &  $\underline{99.4}$  &  $\mathbf{99.4}$  &  $\underline{99.4}$  &  $\mathbf{99.4}$  &  $\mathbf{99.4}$  &  $\underline{99.4}$  &  $99.4$  &  $\underline{99.3}$  &  $\underline{99.3}$\\
\rowcolor{oursrow} \textbf{Ours}~(v3-H+)  &  $\underline{99.4}$  &  $\underline{99.4}$  &  $\underline{99.4}$  &  $99.3$  &  $\mathbf{99.2}$  &  $\mathbf{99.2}$  &  $\underline{99.4}$  &  $\mathbf{99.4}$  &  $\underline{99.4}$  &  $\mathbf{99.4}$  &  $\underline{99.4}$  &  $\mathbf{99.4}$  &  $\mathbf{99.4}$  &  $\underline{99.4}$  &  $99.4$  &  $\mathbf{99.4}$  &  $\mathbf{99.4}$\\
\bottomrule
\end{tabularx}
\endgroup
\end{table*}

\begin{table*}[tb!]
\centering
\caption{\textbf{Performance Comparison on the AIGCDetectionBenchmark Benchmarks.}}
\label{tab:aigc-detection}
\begingroup
\scriptsize
\setlength{\tabcolsep}{1.0pt}
\renewcommand{\arraystretch}{0.84}
\begin{tabularx}{\linewidth}{
    >{\raggedright\arraybackslash}p{1.55cm}
    *{17}{C}
    @{\hspace{1.8pt}}
    !{\color{black!30}\vrule width 0.45pt}
    @{\hspace{1.8pt}}
    C
}
\toprule
\textbf{Method}
& ADM
& \makecell{DALL\\-E2}
& GLIDE
& MJ
& VQDM
& BigGAN
& \makecell{Cycle\\GAN}
& \makecell{Gau\\GAN}
& \makecell{Pro\\GAN}
& SDXL
& SD14
& SD15
& \makecell{Star\\GAN}
& \makecell{Style\\GAN}
& \makecell{Style\\GAN2}
& WFR
& Wukong
& \textbf{Avg.}
\\
\midrule
NPR  &  $43.8$  &  $20.0$  &  $41.2$  &  $53.4$  &  $48.4$  &  $53.1$  &  $76.6$  &  $42.2$  &  $58.7$  &  $59.6$  &  $55.1$  &  $55.0$  &  $67.4$  &  $57.9$  &  $54.6$  &  $58.8$  &  $57.4$  &  $53.1$\\
UnivFD  &  $62.5$  &  $50.0$  &  $61.3$  &  $55.1$  &  $76.9$  &  $87.5$  &  $96.9$  &  $\underline{98.8}$  &  $\mathbf{99.4}$  &  $58.2$  &  $55.6$  &  $55.7$  &  $95.1$  &  $80.0$  &  $69.4$  &  $69.2$  &  $61.1$  &  $72.5$\\
FatFormer  &  $80.2$  &  $68.5$  &  $91.1$  &  $54.4$  &  $88.0$  &  $\mathbf{99.2}$  &  $\mathbf{99.5}$  &  $\mathbf{99.1}$  &  $98.5$  &  $71.7$  &  $67.5$  &  $67.2$  &  $\underline{99.4}$  &  $\mathbf{98.0}$  &  $\mathbf{98.8}$  &  $88.3$  &  $75.6$  &  $85.0$\\
SAFE  &  $49.5$  &  $49.5$  &  $53.0$  &  $49.0$  &  $50.2$  &  $52.2$  &  $51.9$  &  $50.0$  &  $50.0$  &  $49.8$  &  $49.7$  &  $49.8$  &  $50.1$  &  $50.0$  &  $50.0$  &  $49.8$  &  $50.3$  &  $50.3$\\
C2P-CLIP  &  $71.6$  &  $52.3$  &  $73.5$  &  $56.6$  &  $73.7$  &  $\underline{98.4}$  &  $96.8$  &  $\underline{98.8}$  &  $\underline{99.3}$  &  $62.3$  &  $77.5$  &  $76.9$  &  $\mathbf{99.6}$  &  $93.1$  &  $79.4$  &  $94.8$  &  $79.4$  &  $81.4$\\
AIDE  &  $52.9$  &  $51.1$  &  $60.2$  &  $49.8$  &  $69.3$  &  $70.1$  &  $93.6$  &  $60.6$  &  $89.0$  &  $49.6$  &  $51.6$  &  $51.0$  &  $72.1$  &  $66.5$  &  $59.0$  &  $80.6$  &  $54.5$  &  $63.6$\\
DRCT  &  $79.9$  &  $89.2$  &  $89.2$  &  $85.5$  &  $88.6$  &  $81.4$  &  $91.0$  &  $93.8$  &  $71.1$  &  $88.3$  &  $91.4$  &  $91.0$  &  $53.0$  &  $62.7$  &  $63.8$  &  $73.9$  &  $90.8$  &  $81.4$\\
Aligned  &  $51.6$  &  $52.0$  &  $55.6$  &  $96.2$  &  $72.1$  &  $51.2$  &  $49.5$  &  $50.8$  &  $50.7$  &  $95.1$  &  $\mathbf{99.7}$  &  $\mathbf{99.6}$  &  $53.8$  &  $52.7$  &  $51.6$  &  $50.0$  &  $\mathbf{99.6}$  &  $66.6$\\
\midrule
DDA~(v2-L)  &  $89.5$  &  $94.6$  &  $89.6$  &  $95.6$  &  $76.6$  &  $91.0$  &  $72.5$  &  $92.7$  &  $92.8$  &  $\underline{99.4}$  &  $98.7$  &  $98.6$  &  $72.7$  &  $87.8$  &  $90.2$  &  $52.1$  &  $98.8$  &  $87.8$\\
\rowcolor{oursrow} \textbf{Ours}~(v2-L)  &  $95.5$  &  $97.1$  &  $96.7$  &  $95.4$  &  $93.0$  &  $96.0$  &  $94.5$  &  $97.7$  &  $86.0$  &  $98.5$  &  $98.0$  &  $98.0$  &  $75.5$  &  $89.5$  &  $91.6$  &  $94.5$  &  $97.9$  &  $93.8$\\
\addlinespace[1.5pt]
DDA~(v3-L)  &  $96.9$  &  $\underline{99.1}$  &  $97.4$  &  $97.4$  &  $98.2$  &  $95.7$  &  $93.8$  &  $96.6$  &  $92.0$  &  $\underline{99.4}$  &  $98.4$  &  $98.4$  &  $74.1$  &  $89.6$  &  $91.9$  &  $96.2$  &  $98.4$  &  $94.9$\\
\rowcolor{oursrow} \textbf{Ours}~(v3-L)  &  $97.7$  &  $98.4$  &  $98.2$  &  $96.8$  &  $98.5$  &  $97.0$  &  $96.0$  &  $98.0$  &  $92.0$  &  $99.3$  &  $98.7$  &  $98.8$  &  $82.5$  &  $\underline{94.4}$  &  $\underline{94.5}$  &  $\underline{96.3}$  &  $98.8$  &  $96.2$\\
\addlinespace[1.5pt]
DDA~(v3-H+)  &  $\mathbf{98.7}$  &  $\mathbf{99.3}$  &  $\underline{98.8}$  &  $\underline{98.0}$  &  $\mathbf{99.0}$  &  $97.7$  &  $95.8$  &  $97.5$  &  $90.0$  &  $\mathbf{99.5}$  &  $\underline{99.1}$  &  $\underline{99.1}$  &  $94.9$  &  $91.6$  &  $93.0$  &  $89.8$  &  $\underline{99.1}$  &  $\underline{96.5}$\\
\rowcolor{oursrow} \textbf{Ours}~(v3-H+)  &  $\underline{98.6}$  &  $\underline{99.1}$  &  $\mathbf{98.9}$  &  $\mathbf{98.1}$  &  $\underline{98.8}$  &  $97.8$  &  $\underline{97.7}$  &  $\underline{98.8}$  &  $91.2$  &  $\underline{99.4}$  &  $99.0$  &  $\underline{99.1}$  &  $89.1$  &  $93.9$  &  $94.4$  &  $\mathbf{98.4}$  &  $99.0$  &  $\mathbf{97.1}$\\
\bottomrule
\end{tabularx}
\endgroup
\end{table*}

\begin{table*}[tb!]
\centering
\caption{\textbf{Performance Comparison on the ForenSynths Benchmarks.}}
\label{tab:compare-ForenSynths}
\begingroup
\scriptsize
\setlength{\tabcolsep}{1.2pt}
\renewcommand{\arraystretch}{0.84}
\begin{tabularx}{\linewidth}{
    >{\raggedright\arraybackslash}p{1.65cm}
    *{13}{C}
    @{\hspace{1.8pt}}
    !{\color{black!30}\vrule width 0.45pt}
    @{\hspace{1.8pt}}
    C
}
\toprule
\textbf{Method}
& BigGAN & CRN & \makecell{Cycle\\GAN} & \makecell{Deep\\Fake}
& GauGAN & IMLE & ProGAN & SAN & \makecell{Seeing\\Dark}
& \makecell{Star\\GAN} & \makecell{Style\\GAN} & \makecell{Style\\GAN2} & WFR
& \textbf{Avg.}
\\
\midrule
NPR  &  $53.1$  &  $0.4$  &  $76.6$  &  $35.7$  &  $42.2$  &  $5.3$  &  $58.7$  &  $48.4$  &  $63.6$  &  $67.4$  &  $57.9$  &  $54.6$  &  $58.8$  &  $47.9$\\
UnivFD  &  $87.5$  &  $55.7$  &  $96.9$  &  $69.4$  &  $\underline{98.8}$  &  $68.1$  &  $\mathbf{99.4}$  &  $58.2$  &  $62.2$  &  $95.1$  &  $80.0$  &  $69.4$  &  $69.2$  &  $77.7$\\
FatFormer  &  $\mathbf{99.3}$  &  $72.1$  &  $\mathbf{99.5}$  &  $\mathbf{93.0}$  &  $\mathbf{99.3}$  &  $72.1$  &  $98.4$  &  $70.8$  &  $81.9$  &  $\underline{99.4}$  &  $\mathbf{98.1}$  &  $\mathbf{98.9}$  &  $88.3$  &  $90.1$\\
SAFE  &  $52.2$  &  $50.0$  &  $51.9$  &  $50.1$  &  $50.0$  &  $50.0$  &  $50.0$  &  $50.9$  &  $41.1$  &  $50.1$  &  $50.0$  &  $50.0$  &  $49.8$  &  $49.7$\\
C2P-CLIP  &  $\underline{98.4}$  &  $93.3$  &  $96.8$  &  $\underline{92.6}$  &  $\underline{98.8}$  &  $93.2$  &  $\underline{99.3}$  &  $63.2$  &  $\mathbf{94.7}$  &  $\mathbf{99.6}$  &  $93.1$  &  $79.4$  &  $94.8$  &  $\underline{92.1}$\\
AIDE  &  $70.1$  &  $12.2$  &  $93.6$  &  $53.2$  &  $60.6$  &  $15.9$  &  $89.0$  &  $55.3$  &  $44.2$  &  $72.1$  &  $66.5$  &  $59.0$  &  $80.6$  &  $59.4$\\
DRCT  &  $81.4$  &  $78.4$  &  $91.0$  &  $51.5$  &  $93.8$  &  $82.6$  &  $71.1$  &  $84.9$  &  $72.2$  &  $53.0$  &  $62.7$  &  $63.8$  &  $73.9$  &  $73.9$\\
Aligned  &  $51.2$  &  $50.4$  &  $49.5$  &  $71.7$  &  $50.8$  &  $49.7$  &  $50.7$  &  $67.6$  &  $51.4$  &  $53.8$  &  $52.7$  &  $51.6$  &  $50.0$  &  $53.9$\\
\midrule
DDA~(v2-L)  &  $91.0$  &  $87.0$  &  $72.5$  &  $76.5$  &  $92.7$  &  $89.7$  &  $92.8$  &  $94.7$  &  $58.6$  &  $72.7$  &  $87.8$  &  $90.2$  &  $52.1$  &  $81.4$\\
\rowcolor{oursrow} \textbf{Ours}~(v2-L)  &  $96.0$  &  $\underline{94.4}$  &  $94.5$  &  $91.0$  &  $97.7$  &  $\underline{94.6}$  &  $86.0$  &  $\underline{95.9}$  &  $80.3$  &  $75.5$  &  $89.5$  &  $91.6$  &  $94.5$  &  $90.9$\\
\addlinespace[1.5pt]
DDA~(v3-L)  &  $95.7$  &  $66.0$  &  $93.8$  &  $71.2$  &  $96.6$  &  $66.0$  &  $92.0$  &  $93.0$  &  $83.9$  &  $74.1$  &  $89.6$  &  $91.9$  &  $96.2$  &  $85.4$\\
\rowcolor{oursrow} \textbf{Ours}~(v3-L)  &  $97.0$  &  $89.3$  &  $96.0$  &  $86.4$  &  $98.0$  &  $89.3$  &  $92.0$  &  $93.7$  &  $81.9$  &  $82.5$  &  $\underline{94.4}$  &  $\underline{94.5}$  &  $\underline{96.3}$  &  $91.6$\\
\addlinespace[1.5pt]
DDA~(v3-H+)  &  $97.7$  &  $77.9$  &  $95.8$  &  $71.7$  &  $97.5$  &  $77.9$  &  $90.0$  &  $93.9$  &  $78.3$  &  $94.9$  &  $91.6$  &  $93.0$  &  $89.8$  &  $88.5$\\
\rowcolor{oursrow} \textbf{Ours}~(v3-H+)  &  $97.8$  &  $\mathbf{99.2}$  &  $\underline{97.7}$  &  $76.8$  &  $\underline{98.8}$  &  $\mathbf{99.2}$  &  $91.2$  &  $\mathbf{97.0}$  &  $\underline{86.9}$  &  $89.1$  &  $93.9$  &  $94.4$  &  $\mathbf{98.4}$  &  $\mathbf{93.9}$\\
\bottomrule
\end{tabularx}
\endgroup
\end{table*}

\begin{table}[tb!]
  \centering
\caption{\textbf{Performance Comparison on the GenImage Benchmarks.} }
\label{tab:compare-genimage}
\begingroup
\scriptsize
\setlength{\tabcolsep}{1.5pt}
\renewcommand{\arraystretch}{0.84}
\begin{tabularx}{\linewidth}{
    >{\raggedright\arraybackslash}p{1.45cm}
    *{8}{C}
    @{\hspace{1.5pt}}
    !{\color{black!30}\vrule width 0.45pt}
    @{\hspace{1.5pt}}
    C
}
\toprule
\textbf{Method}
& MJ & \makecell{SD\\1.4} & \makecell{SD\\1.5} & ADM
& GLIDE & \makecell{Wu\\kong} & VQDM & \makecell{Big\\GAN}
& \textbf{Avg.}
\\
\midrule
NPR  &  $53.4$  &  $55.1$  &  $55.0$  &  $43.8$  &  $41.2$  &  $57.4$  &  $48.4$  &  $57.7$  &  $51.5$\\
UnivFD  &  $55.1$  &  $55.6$  &  $55.7$  &  $62.5$  &  $61.3$  &  $61.1$  &  $76.9$  &  $84.4$  &  $64.1$\\
FatFormer  &  $52.1$  &  $53.6$  &  $53.8$  &  $61.4$  &  $65.5$  &  $60.9$  &  $72.5$  &  $82.2$  &  $62.8$\\
SAFE  &  $49.0$  &  $49.7$  &  $49.8$  &  $49.5$  &  $53.0$  &  $50.3$  &  $50.2$  &  $50.9$  &  $50.3$\\
C2P-CLIP  &  $56.6$  &  $77.5$  &  $76.9$  &  $71.6$  &  $73.5$  &  $79.4$  &  $73.7$  &  $85.9$  &  $74.4$\\
AIDE  &  $58.2$  &  $77.2$  &  $77.4$  &  $50.4$  &  $54.6$  &  $70.5$  &  $50.8$  &  $50.6$  &  $61.2$\\
DRCT  &  $82.4$  &  $88.3$  &  $88.2$  &  $76.9$  &  $86.1$  &  $87.9$  &  $85.4$  &  $87.0$  &  $85.3$\\
Aligned  &  $97.5$  &  $\mathbf{99.7}$  &  $\mathbf{99.6}$  &  $52.4$  &  $57.6$  &  $\mathbf{99.6}$  &  $75.0$  &  $50.6$  &  $79.0$\\
\midrule
DDA~(v2-L)  &  $95.6$  &  $98.7$  &  $98.6$  &  $89.5$  &  $89.6$  &  $98.7$  &  $76.5$  &  $86.5$  &  $91.7$\\
\rowcolor{oursrow} \textbf{Ours}~(v2-L)  &  $95.4$  &  $98.0$  &  $98.0$  &  $95.5$  &  $96.7$  &  $97.9$  &  $93.0$  &  $97.2$  &  $96.5$\\
\addlinespace[1.5pt]
DDA~(v3-L)  &  $97.4$  &  $98.4$  &  $98.4$  &  $96.9$  &  $97.4$  &  $98.4$  &  $98.2$  &  $96.8$  &  $97.7$\\
\rowcolor{oursrow} \textbf{Ours}~(v3-L)  &  $96.8$  &  $98.7$  &  $98.8$  &  $97.7$  &  $98.2$  &  $98.8$  &  $98.5$  &  $\underline{98.4}$  &  $\underline{98.2}$\\
\addlinespace[1.5pt]
DDA~(v3-H+)  &  $\underline{98.0}$  &  $\underline{99.1}$  &  $\underline{99.1}$  &  $\mathbf{98.7}$  &  $\underline{98.8}$  &  $\underline{99.1}$  &  $\mathbf{99.0}$  &  $\underline{98.4}$  &  $\mathbf{98.8}$\\
\rowcolor{oursrow} \textbf{Ours}~(v3-H+)  &  $\mathbf{98.1}$  &  $99.0$  &  $\underline{99.1}$  &  $\underline{98.6}$  &  $\mathbf{98.9}$  &  $99.0$  &  $\underline{98.8}$  &  $\mathbf{98.7}$  &  $\mathbf{98.8}$\\
\bottomrule
\end{tabularx}
\vspace{-2mm}
\endgroup
\end{table}

\begin{table}[tb!]
\centering
\caption{\textbf{Performance Comparison on the Synthbuster Benchmarks.}}
\label{tab:compare-synthbuster}
\begingroup
\scriptsize
\setlength{\tabcolsep}{1.1pt}
\renewcommand{\arraystretch}{0.84}
\begin{tabularx}{\linewidth}{
    >{\raggedright\arraybackslash}p{1.30cm}
    *{9}{C}
    @{\hspace{1.5pt}}
    !{\color{black!30}\vrule width 0.45pt}
    @{\hspace{1.5pt}}
    C
}
\toprule
\textbf{Method}
& \makecell{DALL\\-E2} & \makecell{DALL\\-E3} & Firefly & GLIDE & MJ
& \makecell{SD\\1.3} & \makecell{SD\\1.4} & SD2 & SDXL
& \textbf{Avg.}
\\
\midrule
NPR  &  $51.1$  &  $49.3$  &  $46.5$  &  $48.5$  &  $52.8$  &  $51.4$  &  $51.8$  &  $46.0$  &  $52.8$  &  $50.0$\\
UnivFD  &  $83.5$  &  $47.4$  &  $89.9$  &  $53.3$  &  $52.5$  &  $70.4$  &  $69.9$  &  $75.7$  &  $68.0$  &  $67.8$\\
FatFormer  &  $59.4$  &  $39.5$  &  $60.3$  &  $72.7$  &  $44.4$  &  $53.7$  &  $54.0$  &  $52.3$  &  $69.1$  &  $56.1$\\
SAFE  &  $58.0$  &  $9.9$  &  $10.3$  &  $52.2$  &  $56.7$  &  $59.4$  &  $59.1$  &  $53.0$  &  $59.5$  &  $46.5$\\
C2P-CLIP  &  $55.6$  &  $63.2$  &  $59.5$  &  $86.7$  &  $52.9$  &  $75.2$  &  $76.7$  &  $69.2$  &  $77.7$  &  $68.5$\\
AIDE  &  $34.9$  &  $33.7$  &  $24.8$  &  $65.0$  &  $57.5$  &  $74.1$  &  $73.7$  &  $53.2$  &  $68.4$  &  $53.9$\\
DRCT  &  $77.2$  &  $86.6$  &  $84.1$  &  $82.6$  &  $73.7$  &  $86.6$  &  $86.6$  &  $83.2$  &  $71.3$  &  $81.3$\\
Aligned  &  $50.2$  &  $48.9$  &  $51.7$  &  $53.5$  &  $\mathbf{98.7}$  &  $\mathbf{98.8}$  &  $\mathbf{98.8}$  &  $\mathbf{98.6}$  &  $\underline{97.3}$  &  $77.4$\\
\midrule
DDA~(v2-L)  &  $86.3$  &  $90.0$  &  $91.9$  &  $76.5$  &  $93.5$  &  $92.9$  &  $92.7$  &  $93.3$  &  $93.5$  &  $90.1$\\
\rowcolor{oursrow} \textbf{Ours}~(v2-L)  &  $\mathbf{98.0}$  &  $\mathbf{97.6}$  &  $\mathbf{98.0}$  &  $94.7$  &  $\underline{98.1}$  &  $\underline{98.0}$  &  $\underline{97.9}$  &  $\underline{98.1}$  &  $\mathbf{98.1}$  &  $\mathbf{97.6}$\\
\addlinespace[1.5pt]
DDA~(v3-L)  &  $90.6$  &  $90.8$  &  $90.8$  &  $88.8$  &  $90.9$  &  $90.9$  &  $90.8$  &  $90.9$  &  $90.9$  &  $90.6$\\
\rowcolor{oursrow} \textbf{Ours}~(v3-L)  &  $93.0$  &  $92.6$  &  $92.9$  &  $92.7$  &  $93.1$  &  $93.1$  &  $93.1$  &  $93.0$  &  $93.1$  &  $92.9$\\
\addlinespace[1.5pt]
\scalebox{0.99}{DDA~(v3-H+)}  &  $\underline{96.8}$  &  $\underline{96.8}$  &  $\underline{96.8}$  &  $\mathbf{96.7}$  &  $96.8$  &  $96.8$  &  $96.8$  &  $96.8$  &  $96.8$  &  $\underline{96.8}$\\
\rowcolor{oursrow} \scalebox{0.99}{\textbf{Ours}~(v3-H+)}  &  $95.8$  &  $95.8$  &  $95.7$  &  $\underline{95.7}$  &  $95.8$  &  $95.8$  &  $95.8$  &  $95.8$  &  $95.8$  &  $95.8$\\
\bottomrule
\end{tabularx}
\endgroup
\end{table}

\begin{table}[tb!]
\centering
\caption{\textbf{Performance Comparison on SynthWildx and WildRF.} }
\label{tab:compare-synthwildx-wildrf}
\begingroup
\scriptsize
\setlength{\tabcolsep}{1.5pt}
\renewcommand{\arraystretch}{0.84}
\begin{tabularx}{\linewidth}{
    >{\raggedright\arraybackslash}p{1.35cm}
    *{4}{C}
    @{\hspace{1.8pt}}
    !{\color{black!30}\vrule width 0.45pt}
    @{\hspace{1.8pt}}
    *{4}{C}
}
\toprule
\multirow{2}{*}{\textbf{Method}}
& \multicolumn{4}{c}{\textbf{SynthWildx}}
& \multicolumn{4}{c}{\textbf{WildRF}}
\\
\cmidrule(lr){2-5} \cmidrule(lr){6-9}
& \makecell{DALL\\-E3} & Firefly & MJ & \textbf{Avg.}
& \makecell{Face\\book} & Reddit & Twitter & \textbf{Avg.}
\\
\midrule
NPR  &  $43.6$  &  $61.3$  &  $44.5$  &  $49.8$  &  $78.1$  &  $61.0$  &  $51.3$  &  $63.5$\\
UnivFD  &  $45.4$  &  $65.3$  &  $46.2$  &  $52.3$  &  $49.1$  &  $60.2$  &  $56.5$  &  $55.3$\\
FatFormer  &  $46.5$  &  $61.6$  &  $48.3$  &  $52.1$  &  $54.1$  &  $68.1$  &  $54.4$  &  $58.9$\\
SAFE  &  $49.4$  &  $48.2$  &  $49.6$  &  $49.1$  &  $50.9$  &  $74.1$  &  $37.5$  &  $54.2$\\
C2P-CLIP  &  $56.9$  &  $61.4$  &  $53.0$  &  $57.1$  &  $54.4$  &  $68.4$  &  $55.9$  &  $59.6$\\
AIDE  &  $63.4$  &  $48.8$  &  $51.9$  &  $54.7$  &  $57.8$  &  $71.5$  &  $45.8$  &  $58.4$\\
DRCT  &  $58.3$  &  $56.4$  &  $50.5$  &  $55.1$  &  $46.6$  &  $53.1$  &  $55.2$  &  $51.6$\\
Aligned  &  $85.5$  &  $58.5$  &  $\underline{92.2}$  &  $78.8$  &  $89.4$  &  $69.1$  &  $81.8$  &  $80.1$\\
\midrule
DDA~(v2-L)  &  $92.3$  &  $87.3$  &  $\mathbf{93.1}$  &  $90.9$  &  $93.1$  &  $86.4$  &  $91.5$  &  $90.3$\\
\rowcolor{oursrow} \textbf{Ours}~(v2-L)  &  $\underline{92.5}$  &  $\underline{91.3}$  &  $91.8$  &  $\underline{91.9}$  &  $92.8$  &  $90.6$  &  $94.8$  &  $92.8$\\
\addlinespace[1.5pt]
DDA~(v3-L)  &  $87.9$  &  $86.3$  &  $87.9$  &  $87.4$  &  $90.3$  &  $91.4$  &  $93.5$  &  $91.7$\\
\rowcolor{oursrow} \textbf{Ours}~(v3-L)  &  $91.9$  &  $90.9$  &  $91.0$  &  $91.3$  &  $\mathbf{95.0}$  &  $\mathbf{96.0}$  &  $\underline{95.7}$  &  $\underline{95.6}$\\
\addlinespace[1.5pt] \scalebox{0.99}{DDA~(v3-H+)}  &  $88.5$  &  $88.0$  &  $88.3$  &  $88.3$  &  $\mathbf{95.0}$  &  $94.1$  &  $91.7$  &  $93.6$\\
\rowcolor{oursrow} \scalebox{0.99}{\textbf{Ours}~(v3-H+)}  &  $\mathbf{92.6}$  &  $\mathbf{92.0}$  &  $92.1$  &  $\mathbf{92.2}$  &  $\underline{93.8}$  &  $\underline{95.8}$  &  $\mathbf{98.1}$  &  $\mathbf{95.9}$\\
\bottomrule
\end{tabularx}
\endgroup
\end{table}

\begin{table}[tb!]
  \centering
  \caption{\textbf{Performance Comparison on the DDA-CoCo Benchmarks.}}
  \label{tab:compare-DDA-coco}
\begingroup
\scriptsize
\setlength{\tabcolsep}{1.5pt}
\renewcommand{\arraystretch}{0.84}
\begin{tabularx}{\linewidth}{
    >{\raggedright\arraybackslash}p{1.35cm}
    C *{6}{C}
    @{\hspace{1.5pt}}
    !{\color{black!30}\vrule width 0.45pt}
    @{\hspace{1.5pt}}
    C
}
\toprule
\textbf{Method}
& \textbf{Real}
& XL & EMA & MSE & SD21 & SD35 & FLUX.1
& \textbf{Avg.}
\\
\midrule
      NPR  &  $55.4$  &  $16.1$  &  $31.1$  &  $41.3$  &  $41.2$  &  $24.9$  &  $19.2$  &  $42.2$\\
      UnivFD  &  $99.2$  &  $3.9$  &  $9.6$  &  $7.3$  &  $7.4$  &  $3.4$  &  $1.8$  &  $52.4$\\
      FatFormer  &  $96.4$  &  $5.4$  &  $6.9$  &  $10.4$  &  $10.3$  &  $6.6$  &  $2.8$  &  $51.7$\\
      SAFE  &  $98.8$  &  $0.6$  &  $0.9$  &  $0.9$  &  $1.0$  &  $0.3$  &  $1.8$  &  $49.9$\\
      C2P-CLIP  &  $\underline{99.5}$  &  $2.0$  &  $2.7$  &  $4.3$  &  $4.2$  &  $4.0$  &  $1.3$  &  $51.3$\\
      AIDE  &  $98.8$  &  $0.6$  &  $2.2$  &  $1.7$  &  $1.8$  &  $0.3$  &  $0.6$  &  $50.0$\\
      DRCT  &  $94.2$  &  $16.9$  &  $34.8$  &  $33.5$  &  $33.6$  &  $21.7$  &  $17.2$  &  $60.2$\\
      Aligned  &  $\mathbf{99.8}$  &  $82.5$  &  $99.2$  &  $99.0$  &  $99.1$  &  $55.4$  &  $3.6$  &  $86.5$\\
      \midrule
DDA~(v2-L)  &  $99.0$  &  $95.0$  &  $99.3$  &  $99.7$  &  $\underline{99.7}$  &  $68.1$  &  $50.2$  &  $92.2$\\
\rowcolor{oursrow} \textbf{Ours}~(v2-L)  &  $97.5$  &  $\underline{99.4}$  &  $\mathbf{99.9}$  &  $\mathbf{100.0}$  &  $\mathbf{99.9}$  &  $86.1$  &  $69.3$  &  $95.0$\\
\addlinespace[1.5pt]
      DDA~(v3-L)  &  $98.3$  &  $97.1$  &  $99.6$  &  $99.7$  &  $\underline{99.7}$  &  $83.6$  &  $85.0$  &  $96.2$\\
\rowcolor{oursrow} \textbf{Ours}~(v3-L)  &  $98.6$  &  $97.3$  &  $99.6$  &  $99.6$  &  $99.6$  &  $\underline{88.4}$  &  $77.5$  &  $96.1$\\
\addlinespace[1.5pt]
      \scalebox{0.99}{DDA~(v3-H+)}  &  $98.9$  &  $99.3$  &  $\underline{99.8}$  &  $\underline{99.9}$  &  $\mathbf{99.9}$  &  $81.4$  &  $\underline{86.5}$  &  $\underline{96.7}$\\
      \rowcolor{oursrow} \scalebox{0.99}{\textbf{Ours}~(v3-H+)}  &  $98.8$  &  $\mathbf{99.6}$  &  $\mathbf{99.9}$  &  $\underline{99.9}$  &  $\mathbf{99.9}$  &  $\mathbf{90.9}$  &  $\mathbf{86.8}$  &  $\mathbf{97.5}$\\
      
\bottomrule
\end{tabularx}
\endgroup
\end{table}

\begin{table}[tb!]
\centering
\caption{\textbf{Performance Comparison on the EvalGEN Benchmarks.}}
\label{tab:compare-eval-gen}

\begingroup
\scriptsize
\setlength{\tabcolsep}{4.6pt}
\renewcommand{\arraystretch}{0.86}

\begin{tabularx}{\linewidth}{
    l
    *{5}{Y}
    !{\color{black!25}\vrule width 0.45pt}
    Y
}
\toprule
\textbf{Method}
& \textbf{Flux}
& \textbf{GoT}
& \textbf{Infinity}
& \textbf{NOVA}
& \textbf{OmiGen}
& \textbf{Avg.}
\\
\midrule

NPR
& $0.7$
& $0.2$
& $6.5$
& $4.7$
& $2.2$
& $2.9$
\\

UnivFD
& $4.0$
& $9.2$
& $15.7$
& $8.3$
& $39.6$
& $15.4$
\\

FatFormer
& $9.9$
& $47.9$
& $44.7$
& $98.3$
& $27.3$
& $45.6$
\\

SAFE
& $1.0$
& $0.5$
& $1.9$
& $0.6$
& $1.6$
& $1.1$
\\

C2P-CLIP
& $8.7$
& $49.6$
& $35.3$
& $86.4$
& $14.5$
& $38.9$
\\

AIDE
& $17.9$
& $24.7$
& $3.4$
& $16.3$
& $33.4$
& $19.1$
\\

DRCT
& $72.5$
& $81.4$
& $77.9$
& $84.6$
& $72.5$
& $77.8$
\\

Aligned
& $32.0$
& $72.3$
& $74.0$
& $84.8$
& $77.0$
& $68.0$
\\

\midrule

DDA~(v2-L)
& $89.9$
& $99.5$
& $97.8$
& $99.5$
& $99.5$
& $97.2$
\\

\rowcolor{oursrow}
\textbf{Ours}~(v2-L)
& $72.1$
& $98.5$
& $\underline{99.3}$
& $\underline{99.9}$
& $98.9$
& $93.8$
\\

\addlinespace[1.8pt]

DDA~(v3-L)
& $\mathbf{97.8}$
& $\mathbf{99.9}$
& $\mathbf{100.0}$
& $\mathbf{100.0}$
& $\mathbf{99.9}$
& $\mathbf{99.5}$
\\

\rowcolor{oursrow}
\textbf{Ours}~(v3-L)
& $90.0$
& $99.5$
& $\mathbf{100.0}$
& $\mathbf{100.0}$
& $99.3$
& $97.8$
\\

\addlinespace[1.8pt]

DDA~(v3-H+)
& $\underline{95.6}$
& $\underline{99.8}$
& $\mathbf{100.0}$
& $\underline{99.9}$
& $\mathbf{99.9}$
& $\underline{99.0}$
\\

\rowcolor{oursrow}
\textbf{Ours}~(v3-H+)
& $94.7$
& $99.7$
& $\mathbf{100.0}$
& $\mathbf{100.0}$
& $\underline{99.8}$
& $98.8$
\\

\bottomrule
\end{tabularx}

\endgroup
\end{table}

\begin{table*}[tb!]
\centering
\caption{\textbf{Local forgery detection Comparison on the BR-GEN Benchmark.} }
\label{tab:local-forgery}
\begingroup
\scriptsize
\setlength{\tabcolsep}{2.5pt}
\renewcommand{\arraystretch}{0.84}
\begin{tabularx}{\linewidth}{
    >{\raggedright\arraybackslash}p{1.65cm}
    *{6}{C}
    @{\hspace{2.0pt}}
    !{\color{black!30}\vrule width 0.45pt}
    @{\hspace{2.0pt}}
    *{6}{C}
    @{\hspace{2.0pt}}
    !{\color{black!30}\vrule width 0.45pt}
    @{\hspace{2.0pt}}
    C
}
\toprule
\multirow{2}{*}{\textbf{Method}}
& \multicolumn{6}{c}{\textbf{Stuff}}
& \multicolumn{6}{c}{\textbf{Background}}
& \multirow{2}{*}{\makecell{\textbf{Total}\\\textbf{Avg.}}}
\\
\cmidrule(lr){2-7} \cmidrule(lr){8-13}
& LaMa & MAT & SDXL & BrushNet & PowerPaint & \textbf{Avg.}
& LaMa & MAT & SDXL & BrushNet & PowerPaint & \textbf{Avg.}
&
\\
\midrule
NPR  &  $56.7$  &  $60.6$  &  $59.3$  &  $62.4$  &  $63.6$  &  $60.5$  &  $47.3$  &  $69.8$  &  $74.6$  &  $78.8$  &  $81.1$  &  $70.3$  &  $65.4$\\
UnivFD  &  $66.2$  &  $69.4$  &  $60.4$  &  $67.1$  &  $60.1$  &  $64.6$  &  $87.2$  &  $90.3$  &  $61.9$  &  $67.3$  &  $67.2$  &  $74.8$  &  $69.7$\\
FatFormer  &  $71.6$  &  $75.1$  &  $64.1$  &  $64.1$  &  $69.8$  &  $68.9$  &  $84.6$  &  $94.3$  &  $69.2$  &  $66.8$  &  $74.1$  &  $77.8$  &  $73.4$\\
SAFE  &  $59.9$  &  $52.9$  &  $54.6$  &  $59.4$  &  $56.9$  &  $56.7$  &  $79.0$  &  $57.2$  &  $64.5$  &  $76.6$  &  $72.3$  &  $69.9$  &  $63.3$\\
C2P-CLIP  &  $67.7$  &  $68.3$  &  $60.5$  &  $64.8$  &  $64.4$  &  $65.1$  &  $87.4$  &  $87.7$  &  $68.1$  &  $71.8$  &  $76.8$  &  $78.4$  &  $71.8$\\
AIDE  &  $59.3$  &  $63.0$  &  $65.1$  &  $59.4$  &  $61.2$  &  $61.6$  &  $71.5$  &  $82.3$  &  $85.3$  &  $82.1$  &  $84.1$  &  $81.1$  &  $71.4$\\
DRCT  &  $64.2$  &  $64.2$  &  $60.1$  &  $62.9$  &  $61.7$  &  $62.6$  &  $76.7$  &  $77.6$  &  $71.0$  &  $69.7$  &  $70.4$  &  $73.1$  &  $67.9$\\
Aligned  &  $55.2$  &  $61.9$  &  $57.3$  &  $71.6$  &  $65.1$  &  $62.2$  &  $51.4$  &  $79.1$  &  $74.1$  &  $90.5$  &  $84.9$  &  $76.0$  &  $69.1$\\
\midrule
DDA~(v2-L)  &  $54.7$  &  $57.8$  &  $61.4$  &  $73.9$  &  $69.1$  &  $63.4$  &  $58.0$  &  $73.2$  &  $84.9$  &  $92.7$  &  $90.0$  &  $79.8$  &  $71.6$\\
\rowcolor{oursrow} \textbf{Ours}~(v2-L)  &  $82.8$  &  $86.0$  &  $74.3$  &  $85.4$  &  $80.7$  &  $81.8$  &  $91.1$  &  $94.7$  &  $91.6$  &  $95.3$  &  $93.7$  &  $93.3$  &  $87.6$\\
\addlinespace[1.5pt]
DDA~(v3-L)  &  $66.0$  &  $71.6$  &  $71.9$  &  $79.9$  &  $77.4$  &  $73.4$  &  $78.4$  &  $91.9$  &  $93.4$  &  $95.1$  &  $94.4$  &  $90.6$  &  $82.0$\\
\rowcolor{oursrow} \textbf{Ours}~(v3-L)  &  $\underline{85.6}$  &  $\underline{87.1}$  &  $\underline{81.9}$  &  $87.2$  &  $\underline{84.9}$  &  $\underline{85.3}$  &  $\underline{94.4}$  &  $\underline{96.5}$  &  $\underline{95.4}$  &  $96.5$  &  $\underline{95.8}$  &  $\underline{95.7}$  &  $\underline{90.5}$\\
\addlinespace[1.5pt]
DDA~(v3-H+)  &  $69.9$  &  $77.8$  &  $78.0$  &  $\underline{87.3}$  &  $84.0$  &  $79.4$  &  $86.6$  &  $94.7$  &  $94.9$  &  $\underline{96.7}$  &  $95.4$  &  $93.7$  &  $86.6$\\
\rowcolor{oursrow} \textbf{Ours}~(v3-H+)  &  $\mathbf{90.6}$  &  $\mathbf{92.4}$  &  $\mathbf{87.6}$  &  $\mathbf{92.1}$  &  $\mathbf{90.0}$  &  $\mathbf{90.5}$  &  $\mathbf{96.2}$  &  $\mathbf{97.6}$  &  $\mathbf{96.5}$  &  $\mathbf{97.2}$  &  $\mathbf{96.4}$  &  $\mathbf{96.8}$  &  $\mathbf{93.7}$\\
\bottomrule
\end{tabularx}
\endgroup
\end{table*}

\textbf{Cross-Domain Forgeries.} Table~\ref{tab:cross-forgery} uses GAN- and diffusion-based generators as a diagnostic split for artifact-family balance. Prior methods often show artifact-family specialization: FatFormer and C2P-CLIP achieve strong GAN averages but drop sharply on diffusion generators, whereas reconstruction-aligned baselines such as Aligned and DDA~(v2-L) are more diffusion-oriented. This imbalance is exactly the artifact-bias problem discussed earlier: learning from a narrow aligned artifact source can improve one generator family while leaving another underrepresented. Under the same DINO-v2-L backbone, Ours~(v2-L) raises GAN Avg. from 72.5\% to 89.8\% and Diff. Avg. from 90.0\% to 92.2\% over DDA~(v2-L), improving Avg. Acc. from 81.3\% to 91.0\%. More importantly, the GAN--diffusion gap is reduced from 17.5 points for DDA~(v2-L) to 2.4 points for Ours~(v2-L), showing that the gain is not obtained by overfitting to one side. The same trend remains with stronger backbones: Ours~(v3-L) reaches 94.6\% Avg. Acc., and Ours~(v3-H+) reaches 95.5\%, with 95.1\% GAN Avg. and 95.8\% Diff. Avg. These results support the role of complementary aligned artifacts in building a more balanced decision boundary. Fig.~\ref{fig:tsne} further compares DDA and Ours using the same DINO-v2-L backbone. DDA shows substantial real/fake overlap on CycleGAN, ADM, and VQDM, whereas Ours forms clearer boundaries across both GAN- and diffusion-based generators. This separation agrees with Table~\ref{tab:cross-forgery} and confirms that complementary artifacts improve feature discriminability.

\textbf{Cross-Benchmark Generalization.} Table~\ref{tab:compare-methods} tests transfer across 11 benchmarks with different data sources, generator families, and post-processing histories. Ours~(v2-L) reaches 93.3\% Avg., Ours~(v3-L) reaches 94.3\%, and Ours~(v3-H+) reaches 96.1\%, exceeding DDA~(v3-H+) by 1.4 points. The improvement is especially clear on benchmarks with broader source variation, including ForenSynths, SynthWildx, WildRF, and BFree-Online, where detectors must generalize beyond a single curated generator set. For example, C2P-CLIP reaches 92.1\% on ForenSynths but only 57.1\% and 59.6\% on SynthWildx and WildRF, while DRCT decreases from 90.5\% on DRCT-2M to 55.1\% and 51.6\% on these two in-the-wild benchmarks. The detailed breakdowns in Tables~\ref{tab:compare-drct}--\ref{tab:compare-eval-gen} further show that DRCT-2M, GenImage, DDA-COCO, and EvalGEN are already close to saturation for strong DINO backbones, so their main value is checking whether performance is preserved across many generator variants. This pattern is consistent with our motivation: aligned training reduces dataset shortcuts, but broad artifact coverage is still needed for stable cross-benchmark transfer.

\textbf{Local Forgery Detection.} Table~\ref{tab:local-forgery} evaluates BR-GEN, where forged evidence is spatially limited and can be overwhelmed by real surrounding regions. Ours improves over DDA at every backbone scale, raising Total Avg. from 71.6\% to 87.6\% with v2-L, from 82.0\% to 90.5\% with v3-L, and from 86.6\% to 93.7\% with v3-H+. The gains are strongest on Stuff regions, where object- and texture-level edits introduce subtler traces than background inpainting: the Stuff Avg. improves from 63.4\% to 81.8\% with v2-L and from 79.4\% to 90.5\% with v3-H+. Background regions also improve, but the smaller gap at larger backbone scales suggests that they are less challenging once global context is reliable. These results indicate that the proposed training strategy improves sensitivity to localized forensic cues rather than relying only on image-level source statistics.

\subsection{Ablation Study}

\begin{figure}[!t]
\centering
\includegraphics[width=0.95\columnwidth]{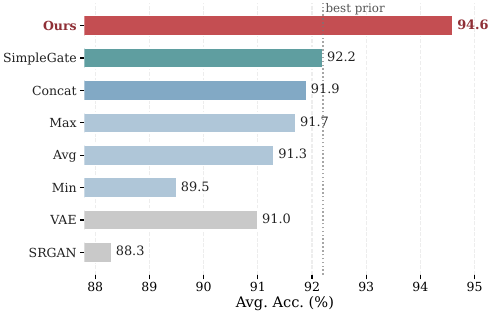}
\caption{\textbf{Ablation on expert fusion strategies.} Avg. Acc. (\%) is reported on the cross-domain forgery benchmark with DINO-v3-L.}
\label{tab:ablation-fusion-strategy}
\end{figure}

All ablation experiments are conducted on the cross-domain forgery benchmark with DINO-v3-L. Fig.~\ref{tab:ablation-fusion-strategy} first compares expert fusion strategies. The single SRGAN and VAE experts obtain 88.3\% and 91.0\% Avg. Acc., respectively, indicating that either artifact source alone provides useful but incomplete evidence. Static feature-level fusion is also limited: min fusion drops to 89.5\%, while average, max, and concatenation reach 91.3\%, 91.7\%, and 91.9\%. SimpleGate improves the result to 92.2\%, showing that adaptive expert weighting is beneficial, but its final-descriptor fusion still misses intermediate forensic cues. LACF achieves 94.6\%, giving a 2.4\% gain over SimpleGate and confirming the importance of layer-wise complementary feature integration.

\begin{figure}[!t]
\centering
\includegraphics[width=0.95\columnwidth]{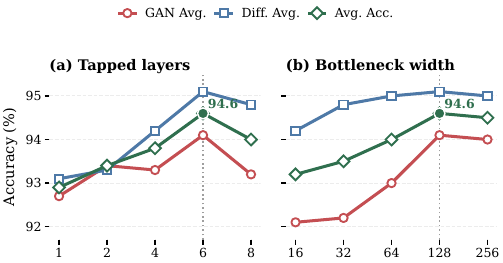}
\caption{\textbf{LACF capacity ablation.} Subplots vary the number of tapped layers $M_{\mathrm{tap}}$ and bottleneck width $d_b$ on the cross-domain forgery benchmark with DINO-v3-L. Values are balanced accuracy (\%).}
\label{tab:ablation-taps}
\label{tab:ablation-bottleneck}
\end{figure}

Fig.~\ref{tab:ablation-taps} studies the capacity of LACF. Increasing the number of tapped layers generally improves aggregate performance, confirming that complementary artifacts are distributed across Transformer depths rather than concentrated only in the final representation. Performance peaks at $M_{\mathrm{tap}}=6$ and then slightly declines with denser tapping, indicating that excessive intermediate features can introduce redundant evidence. Bottleneck width exhibits a similar trend, with performance improving as fusion capacity grows and saturating beyond $d_b=128$. We therefore use $M_{\mathrm{tap}}=6$ and $d_b=128$, which yield the best Avg. Acc. of 94.6\%.

Tables~\ref{tab:ablation-loss} and~\ref{tab:ablation-branches} analyze the fusion objective and branch composition. Using balanced BCE alone reaches 93.2\% Avg. Acc. Adding grouped contrastive learning raises it to 93.9\%, showing that source-aware feature separation helps when VAE and SRGAN fakes should not be compressed into one homogeneous fake cluster. The boundary loss alone gives a smaller gain to 93.4\%, but combining BCE, GCL, and BD achieves the best 94.6\%, indicating that class balance, grouped separation, and real-manifold regularization are complementary. For branch composition, the cross branch gives 92.7\%, adding the SRGAN-oriented or VAE-oriented branch improves the result to 93.1\% and 93.5\%, and using all three branches reaches 94.6\%. This supports the LACF design: cross-artifact interaction is useful, but preserving source-oriented evidence is necessary for the strongest fused representation.

\begin{table}[!t]
\centering
\caption{
\textbf{Ablations on the fusion objective and branch composition.}
(a) evaluates balanced binary cross-entropy (BCE), grouped contrastive
loss (GCL), and real-manifold boundary loss (BD).
(b) evaluates the cross-artifact, VAE-oriented, and SRGAN-oriented
fusion branches.
}
\label{tab:ablation-fusion}

\begingroup
\scriptsize
\setlength{\tabcolsep}{2.5pt}
\renewcommand{\arraystretch}{0.88}
\setlength{\aboverulesep}{0.45ex}
\setlength{\belowrulesep}{0.45ex}

\begin{tabular*}{\columnwidth}{
    @{}
    p{0.475\columnwidth}
    @{\hspace{0.05\columnwidth}}
    p{0.475\columnwidth}
    @{}
}

\begin{minipage}[t]{\linewidth}
\vspace{0pt}
\centering

\refstepcounter{subtable}
\label{tab:ablation-loss}

\textbf{(a) Fusion objectives}
\vspace{2pt}

\begin{tabular*}{\linewidth}{
    @{\extracolsep{\fill}}
    c
    c
    c
    @{\hspace{5pt}}
    c
    @{}
}
\toprule
\textbf{BCE}
& \textbf{GCL}
& \textbf{BD}
& \textbf{Avg.}
\\
\midrule

\checkmark
& \textemdash
& \textemdash
& $93.2$
\\

\checkmark
& \checkmark
& \textemdash
& $93.9$
\\

\checkmark
& \textemdash
& \checkmark
& $93.4$
\\

\addlinespace[1.5pt]

\checkmark
& \checkmark
& \checkmark
& $\mathbf{94.6}$
\\

\bottomrule
\end{tabular*}
\end{minipage}

&

\begin{minipage}[t]{\linewidth}
\vspace{0pt}
\centering

\refstepcounter{subtable}
\label{tab:ablation-branches}

\textbf{(b) Fusion branches}
\vspace{2pt}

\begin{tabular*}{\linewidth}{
    @{\extracolsep{\fill}}
    c
    c
    c
    @{\hspace{5pt}}
    c
    @{}
}
\toprule
\textbf{Cross}
& \textbf{VAE}
& \textbf{SRGAN}
& \textbf{Avg.}
\\
\midrule

\checkmark
& \textemdash
& \textemdash
& $92.7$
\\

\checkmark
& \textemdash
& \checkmark
& $93.1$
\\

\checkmark
& \checkmark
& \textemdash
& $93.5$
\\

\addlinespace[1.5pt]

\checkmark
& \checkmark
& \checkmark
& $\mathbf{94.6}$
\\

\bottomrule
\end{tabular*}
\end{minipage}

\end{tabular*}
\endgroup

\vspace{-2pt}
\end{table}

\begin{table}[t]
\centering
\caption{\textbf{Robustness on the cross-domain forgery benchmark.} Balanced accuracy (\%) is reported under common image perturbations. Ours uses DINO-v2-L as backbone.}
\label{tab:robust}
\begingroup
\scriptsize
\setlength{\tabcolsep}{4.2pt}
\renewcommand{\arraystretch}{0.84}
\begin{tabularx}{\linewidth}{l *{5}{Y}}
\toprule
\textbf{Method} & \textbf{Blur} & \textbf{Crop} & \textbf{JPEG} & \textbf{Noise} & \textbf{Combined} \\
\midrule
SAFE      & $81.4$ & $85.8$ & $70.1$ & $69.9$ & $76.6$ \\
NPR       & $80.2$ & $85.5$ & $74.1$ & $73.0$ & $78.2$ \\
FatFormer & $78.0$ & $81.3$ & $78.4$ & $73.8$ & $77.9$ \\
C2P-CLIP  & $79.2$ & $80.0$ & $74.0$ & $70.9$ & $76.1$ \\
AIDE      & $\underline{82.5}$ & $\underline{86.7}$ & $71.2$ & $\underline{80.1}$ & $\underline{79.9}$ \\
\midrule
DDA       & $79.6$ & $79.9$ & $\underline{80.4}$ & $76.7$ & $79.2$ \\
\rowcolor{oursrow}
\textbf{Ours}~(v2-L) & $\mathbf{85.2}$ & $\mathbf{90.2}$ & $\mathbf{88.5}$ & $\mathbf{85.6}$ & $\mathbf{87.5}$ \\
\bottomrule
\end{tabularx}
\endgroup
\end{table}

\subsection{Robust Analysis}

Following \cite{frank2020leveraging}, we evaluate Ours~(v2-L) under four inference-time perturbations and their combination on the cross-domain forgery benchmark. Each perturbation is applied with 50\% probability: Gaussian blur uses kernels from $\{3,5,7,9\}$, cropping sample ratios from $U(5,20)$ before resizing, JPEG uses quality factors from $U(10,75)$, and Gaussian noise uses variances from $U(5.0,20.0)$. The Combined setting combines all four perturbation types.

As shown in Tab.~\ref{tab:robust}, Ours~(v2-L) ranks first under every perturbation, reaching 85.2\%, 90.2\%, 88.5\%, 85.6\%, and 87.5\% under blur, cropping, JPEG, noise, and Combined, respectively. These results exceed the strongest competitor in each setting by 2.7\%, 3.5\%, 8.1\%, 5.5\%, and 7.6\%, while the five-setting average of 87.4\% is 7.3\% above the best baseline average. The consistent gains across these perturbations, indicate that LACF preserves complementary artifact evidence rather than relying on a single fragile cue.

\section{Conclusion}

In this work, we identify \textbf{artifact bias} as a remaining limitation of aligned training: although content, size, and format alignment suppress dataset shortcuts, a single reconstruction source covers only a narrow artifact manifold. We use SRGAN-based adversarial restoration as a controlled proxy for an artifact-forming mechanism complementary to VAE reconstruction, constructing mechanism-diverse negatives without sacrificing alignment. We further show that directly mixing these heterogeneous sources causes optimization conflict. Accordingly, Artifact-Complementary Expert Fusion (ACEF) trains artifact-specific experts independently, freezes them, and uses Layer-wise Artifact-Complementary Fusion to integrate source-oriented and cross-artifact evidence across Transformer depths. Experiments across 13 benchmarks demonstrate strong and balanced generalization, showing that broader artifact coverage and decoupled fusion are both important for generalizable AIGC detection.

\bibliography{ref}

@inproceedings{ledig2017photo,
  title={Photo-realistic single image super-resolution using a generative adversarial network},
  author={Ledig, Christian and Theis, Lucas and Husz{\'a}r, Ferenc and Caballero, Jose and Cunningham, Andrew and Acosta, Alejandro and Aitken, Andrew and Tejani, Alykhan and Totz, Johannes and Wang, Zehan and others},
  booktitle={Proceedings of the IEEE Conference on Computer Vision and Pattern Recognition},
  pages={4681--4690},
  year={2017}
}

@article{rajan2024aligned,
  title={Aligned datasets improve detection of latent diffusion-generated images},
  author={Rajan, Anirudh Sundara and Ojha, Utkarsh and Schloesser, Jedidiah and Lee, Yong Jae},
  journal={arXiv preprint arXiv:2410.11835},
  year={2024}
}

@inproceedings{guillaro2025bias,
  title={A bias-free training paradigm for more general ai-generated image detection},
  author={Guillaro, Fabrizio and Zingarini, Giada and Usman, Ben and Sud, Avneesh and Cozzolino, Davide and Verdoliva, Luisa},
  booktitle={Proceedings of the IEEE/CVF Conference on Computer Vision and Pattern Recognition},
  pages={18685--18694},
  year={2025}
}

@article{chen2025dual,
  title={Dual data alignment makes ai-generated image detector easier generalizable},
  author={Chen, Ruoxin and Xi, Junwei and Yan, Zhiyuan and Zhang, Ke-Yue and Wu, Shuang and Xie, Jingyi and Chen, Xu and Xu, Lei and Guan, Isabel and Yao, Taiping and others},
  journal={arXiv preprint arXiv:2505.14359},
  year={2025}
}

@inproceedings{chen2024drct,
  title={Drct: Diffusion reconstruction contrastive training towards universal detection of diffusion generated images},
  author={Chen, Baoying and Zeng, Jishen and Yang, Jianquan and Yang, Rui},
  booktitle={International Conference on Machine Learning},
  pages={7621--7639},
  year={2024}
}

@article{liu2025beyond,
  title={Beyond Artifacts: Real-Centric Envelope Modeling for Reliable AI-Generated Image Detection},
  author={Liu, Ruiqi and Han, Yi and Zhang, Zhengbo and Yao, Liwei and Yan, Zhiyuan and Shen, Jialiang and Chen, ZhiJin and Sun, Boyi and Weng, Lubin and Dong, Jing and others},
  journal={arXiv preprint arXiv:2512.20937},
  year={2025}
}

@article{song2020denoising,
  title={Denoising diffusion implicit models},
  author={Song, Jiaming and Meng, Chenlin and Ermon, Stefano},
  journal={arXiv preprint arXiv:2010.02502},
  year={2020}
}

@inproceedings{rombach2022high,
  title={High-resolution image synthesis with latent diffusion models},
  author={Rombach, Robin and Blattmann, Andreas and Lorenz, Dominik and Esser, Patrick and Ommer, Bj{\"o}rn},
  booktitle={Proceedings of the IEEE/CVF Conference on Computer Vision and Pattern Recognition},
  pages={10684--10695},
  year={2022}
}

@article{goodfellow2014generative,
  title={Generative adversarial nets},
  author={Goodfellow, Ian J and Pouget-Abadie, Jean and Mirza, Mehdi and Xu, Bing and Warde-Farley, David and Ozair, Sherjil and Courville, Aaron and Bengio, Yoshua},
  journal={Advances in Neural Information Processing Systems},
  volume={27},
  year={2014}
}

@article{ho2020denoising,
  title={Denoising diffusion probabilistic models},
  author={Ho, Jonathan and Jain, Ajay and Abbeel, Pieter},
  journal={Advances in Neural Information Processing Systems},
  volume={33},
  pages={6840--6851},
  year={2020}
}

@inproceedings{van2016pixel,
  title={Pixel recurrent neural networks},
  author={Van Den Oord, A{\"a}ron and Kalchbrenner, Nal and Kavukcuoglu, Koray},
  booktitle={International Conference on Machine Learning},
  pages={1747--1756},
  year={2016},
  organization={PMLR}
}

@inproceedings{ojha2023towards,
  title={Towards universal fake image detectors that generalize across generative models},
  author={Ojha, Utkarsh and Li, Yuheng and Lee, Yong Jae},
  booktitle={Proceedings of the IEEE/CVF Conference on Computer Vision and Pattern Recognition},
  pages={24480--24489},
  year={2023}
}

@inproceedings{li2025towards,
  title={Towards Generalizable AI-Generated Image Detection via Image-Adaptive Prompt Learning},
  author={Li, Yiheng and Tan, Zichang and Xu, Guoqing and Lei, Zhen and Zhou, Xu and Yang, Yang},
  booktitle={Proceedings of the IEEE/CVF Conference on Computer Vision and Pattern Recognition},
  pages={21262--21272},
  year={2026}
}

@inproceedings{tan2024rethinking,
  title={Rethinking the up-sampling operations in cnn-based generative network for generalizable deepfake detection},
  author={Tan, Chuangchuang and Zhao, Yao and Wei, Shikui and Gu, Guanghua and Liu, Ping and Wei, Yunchao},
  booktitle={Proceedings of the IEEE/CVF Conference on Computer Vision and Pattern Recognition},
  pages={28130--28139},
  year={2024}
}

@inproceedings{tan2023learning,
  title={Learning on gradients: Generalized artifacts representation for gan-generated images detection},
  author={Tan, Chuangchuang and Zhao, Yao and Wei, Shikui and Gu, Guanghua and Wei, Yunchao},
  booktitle={Proceedings of the IEEE/CVF Conference on Computer Vision and Pattern Recognition},
  pages={12105--12114},
  year={2023}
}

@inproceedings{liu2024forgery,
  title={Forgery-aware adaptive transformer for generalizable synthetic image detection},
  author={Liu, Huan and Tan, Zichang and Tan, Chuangchuang and Wei, Yunchao and Wang, Jingdong and Zhao, Yao},
  booktitle={Proceedings of the IEEE/CVF Conference on Computer Vision and Pattern Recognition},
  pages={10770--10780},
  year={2024}
}

@inproceedings{tan2024frequency,
  title={Frequency-aware deepfake detection: Improving generalizability through frequency space domain learning},
  author={Tan, Chuangchuang and Zhao, Yao and Wei, Shikui and Gu, Guanghua and Liu, Ping and Wei, Yunchao},
  booktitle={Proceedings of the AAAI Conference on Artificial Intelligence},
  volume={38},
  number={5},
  pages={5052--5060},
  year={2024}
}

@inproceedings{frank2020leveraging,
  title={Leveraging frequency analysis for deep fake image recognition},
  author={Frank, Joel and Eisenhofer, Thorsten and Sch{\"o}nherr, Lea and Fischer, Asja and Kolossa, Dorothea and Holz, Thorsten},
  booktitle={International Conference on Machine Learning},
  pages={3247--3258},
  year={2020},
  organization={PMLR}
}

@article{odena2016deconvolution,
  title={Deconvolution and checkerboard artifacts},
  author={Odena, Augustus and Dumoulin, Vincent and Olah, Chris},
  journal={Distill},
  volume={1},
  number={10},
  pages={e3},
  year={2016}
}

@inproceedings{durall2020watch,
  title={Watch your up-convolution: Cnn based generative deep neural networks are failing to reproduce spectral distributions},
  author={Durall, Ricard and Keuper, Margret and Keuper, Janis},
  booktitle={Proceedings of the IEEE/CVF Conference on Computer Vision and Pattern Recognition},
  pages={7890--7899},
  year={2020}
}

@inproceedings{wang2020cnn,
  title={CNN-generated images are surprisingly easy to spot... for now},
  author={Wang, Sheng-Yu and Wang, Oliver and Zhang, Richard and Owens, Andrew and Efros, Alexei A},
  booktitle={Proceedings of the IEEE/CVF Conference on Computer Vision and Pattern Recognition},
  pages={8695--8704},
  year={2020}
}

@article{liang2025ferretnet,
  title={FerretNet: Efficient Synthetic Image Detection via Local Pixel Dependencies},
  author={Liang, Shuqiao and Liu, Jian and Chen, Renzhang and Guan, Quanlong},
  journal={arXiv preprint arXiv:2509.20890},
  year={2025}
}

@inproceedings{jia2025secret,
  title={Secret lies in color: Enhancing ai-generated images detection with color distribution analysis},
  author={Jia, Zexi and Huang, Chuanwei and Zhu, Yeshuang and Fei, Hongyan and Duan, Xiaoyue and Yuan, Zhiqiang and Deng, Ying and Zhang, Jiapei and Zhang, Jinchao and Zhou, Jie},
  booktitle={Proceedings of the IEEE/CVF Conference on Computer Vision and Pattern Recognition},
  pages={13445--13454},
  year={2025}
}

@inproceedings{li2025improving,
  title={Improving synthetic image detection towards generalization: An image transformation perspective},
  author={Li, Ouxiang and Cai, Jiayin and Hao, Yanbin and Jiang, Xiaolong and Hu, Yao and Feng, Fuli},
  booktitle={Proceedings of the ACM SIGKDD Conference on Knowledge Discovery and Data Mining},
  pages={2405--2414},
  year={2025}
}

@inproceedings{wang2023dire,
  title={Dire for diffusion-generated image detection},
  author={Wang, Zhendong and Bao, Jianmin and Zhou, Wengang and Wang, Weilun and Hu, Hezhen and Chen, Hong and Li, Houqiang},
  booktitle={Proceedings of the IEEE/CVF International Conference on Computer Vision},
  pages={22445--22455},
  year={2023}
}

@inproceedings{chu2025fire,
  title={Fire: Robust detection of diffusion-generated images via frequency-guided reconstruction error},
  author={Chu, Beilin and Xu, Xuan and Wang, Xin and Zhang, Yufei and You, Weike and Zhou, Linna},
  booktitle={Proceedings of the IEEE/CVF Conference on Computer Vision and Pattern Recognition},
  pages={12830--12839},
  year={2025}
}

@inproceedings{cheng2025co,
  title={Co-spy: Combining semantic and pixel features to detect synthetic images by ai},
  author={Cheng, Siyuan and Lyu, Lingjuan and Wang, Zhenting and Zhang, Xiangyu and Sehwag, Vikash},
  booktitle={Proceedings of the IEEE/CVF Conference on Computer Vision and Pattern Recognition},
  pages={13455--13465},
  year={2025}
}

@inproceedings{luo2024lare,
  title={Lare\^{} 2: Latent reconstruction error based method for diffusion-generated image detection},
  author={Luo, Yunpeng and Du, Junlong and Yan, Ke and Ding, Shouhong},
  booktitle={Proceedings of the IEEE/CVF Conference on Computer Vision and Pattern Recognition},
  pages={17006--17015},
  year={2024}
}

@inproceedings{radford2021learning,
  title={Learning transferable visual models from natural language supervision},
  author={Radford, Alec and Kim, Jong Wook and Hallacy, Chris and Ramesh, Aditya and Goh, Gabriel and Agarwal, Sandhini and Sastry, Girish and Askell, Amanda and Mishkin, Pamela and Clark, Jack and others},
  booktitle={International Conference on Machine Learning},
  pages={8748--8763},
  year={2021},
  organization={PMLR}
}

@inproceedings{caron2021emerging,
  title={Emerging properties in self-supervised vision transformers},
  author={Caron, Mathilde and Touvron, Hugo and Misra, Ishan and J{\'e}gou, Herv{\'e} and Mairal, Julien and Bojanowski, Piotr and Joulin, Armand},
  booktitle={Proceedings of the IEEE/CVF International Conference on Computer Vision},
  pages={9650--9660},
  year={2021}
}

@article{oquab2024dinov2,
  title={DINOv2: Learning robust visual features without supervision},
  author={Oquab, Maxime and Darcet, Timoth{\'e}e and Moutakanni, Th{\'e}o and Vo, Huy V and Szafraniec, Marc and Khalidov, Vasil and Fernandez, Pierre and Haziza, Daniel and Massa, Francisco and El-Nouby, Alaaeldin and others},
  journal={Transactions on Machine Learning Research},
  year={2024}
}

@inproceedings{tan2025c2p,
  title={C2p-clip: Injecting category common prompt in clip to enhance generalization in deepfake detection},
  author={Tan, Chuangchuang and Tao, Renshuai and Liu, Huan and Gu, Guanghua and Wu, Baoyuan and Zhao, Yao and Wei, Yunchao},
  booktitle={Proceedings of the AAAI Conference on Artificial Intelligence},
  volume={39},
  number={7},
  pages={7184--7192},
  year={2025}
}

@article{yan2024orthogonal,
  title={Orthogonal subspace decomposition for generalizable ai-generated image detection},
  author={Yan, Zhiyuan and Wang, Jiangming and Jin, Peng and Zhang, Ke-Yue and Liu, Chengchun and Chen, Shen and Yao, Taiping and Ding, Shouhong and Wu, Baoyuan and Yuan, Li},
  journal={arXiv preprint arXiv:2411.15633},
  year={2024}
}

@article{yan2024sanity,
  title={A sanity check for ai-generated image detection},
  author={Yan, Shilin and Li, Ouxiang and Cai, Jiayin and Hao, Yanbin and Jiang, Xiaolong and Hu, Yao and Xie, Weidi},
  journal={arXiv preprint arXiv:2406.19435},
  year={2024}
}

@article{zhou2026simplicity,
  title={Simplicity Prevails: The Emergence of Generalizable AIGI Detection in Visual Foundation Models},
  author={Zhou, Yue and He, Xinan and Lin, Kaiqing and Fan, Bing and Ding, Feng and Li, Bin},
  journal={arXiv preprint arXiv:2602.01738},
  year={2026}
}

@inproceedings{sha2023fake,
  title={De-fake: Detection and attribution of fake images generated by text-to-image generation models},
  author={Sha, Zeyang and Li, Zheng and Yu, Ning and Zhang, Yang},
  booktitle={Proceedings of the ACM SIGSAC Conference on Computer and Communications Security},
  pages={3418--3432},
  year={2023}
}

@inproceedings{grommelt2024fake,
  title={Fake or jpeg? revealing common biases in generated image detection datasets},
  author={Grommelt, Patrick and Weiss, Louis and Pfreundt, Franz-Josef and Keuper, Janis},
  booktitle={European Conference on Computer Vision},
  pages={80--95},
  year={2024},
  organization={Springer}
}

@inproceedings{yu2024semgir,
  title={SemGIR: Semantic-guided image regeneration based method for AI-generated image detection and attribution},
  author={Yu, Xiao and Chen, Kejiang and Zeng, Kai and Fang, Han and Yang, Zijin and Shang, Xiuwei and Qi, Yuang and Zhang, Weiming and Yu, Nenghai},
  booktitle={Proceedings of the ACM International Conference on Multimedia},
  pages={8480--8488},
  year={2024}
}

@article{simeoni2025dinov3,
  title={Dinov3},
  author={Sim{\'e}oni, Oriane and Vo, Huy V and Seitzer, Maximilian and Baldassarre, Federico and Oquab, Maxime and Jose, Cijo and Khalidov, Vasil and Szafraniec, Marc and Yi, Seungeun and Ramamonjisoa, Micha{\"e}l and others},
  journal={arXiv preprint arXiv:2508.10104},
  year={2025}
}

@article{hu2021lora,
  title={Lora: Low-rank adaptation of large language models},
  author={Hu, Edward J and Shen, Yelong and Wallis, Phillip and Allen-Zhu, Zeyuan and Li, Yuanzhi and Wang, Shean and Wang, Lu and Chen, Weizhu},
  journal={arXiv preprint arXiv:2106.09685},
  year={2021}
}

@inproceedings{cai2026zooming,
  title={Zooming in on fakes: A novel dataset for localized AI-generated image detection with forgery amplification approach},
  author={Cai, Lvpan and Wang, Haowei and Ji, Jiayi and Zhoumen, Yanshu and Chen, Shen and Yao, Taiping and Sun, Xiaoshuai},
  booktitle={Proceedings of the AAAI Conference on Artificial Intelligence},
  volume={40},
  number={4},
  pages={2534--2542},
  year={2026}
}

@inproceedings{lin2014microsoft,
  title={Microsoft coco: Common objects in context},
  author={Lin, Tsung-Yi and Maire, Michael and Belongie, Serge and Hays, James and Perona, Pietro and Ramanan, Deva and Doll{\'a}r, Piotr and Zitnick, C Lawrence},
  booktitle={European Conference on Computer Vision},
  pages={740--755},
  year={2014},
  organization={Springer}
}

@article{zhu2023genimage,
  title={Genimage: A million-scale benchmark for detecting ai-generated image},
  author={Zhu, Mingjian and Chen, Hanting and Yan, Qiangyu and Huang, Xudong and Lin, Guanyu and Li, Wei and Tu, Zhijun and Hu, Hailin and Hu, Jie and Wang, Yunhe},
  journal={Advances in Neural Information Processing Systems},
  volume={36},
  pages={77771--77782},
  year={2023}
}

@article{bammey2023synthbuster,
  title={Synthbuster: Towards detection of diffusion model generated images},
  author={Bammey, Quentin},
  journal={IEEE Open Journal of Signal Processing},
  volume={5},
  pages={1--9},
  year={2024},
  publisher={IEEE}
}

@article{zhong2023patchcraft,
  title={Patchcraft: Exploring texture patch for efficient ai-generated image detection},
  author={Zhong, Nan and Xu, Yiran and Li, Sheng and Qian, Zhenxing and Zhang, Xinpeng},
  journal={arXiv preprint arXiv:2311.12397},
  year={2023}
}

@article{cavia2024real,
  title={Real-time deepfake detection in the real-world},
  author={Cavia, Bar and Horwitz, Eliahu and Reiss, Tal and Hoshen, Yedid},
  journal={arXiv preprint arXiv:2406.09398},
  year={2024}
}

@inproceedings{cozzolino2024raising,
  title={Raising the bar of ai-generated image detection with clip},
  author={Cozzolino, Davide and Poggi, Giovanni and Corvi, Riccardo and Nie{\ss}ner, Matthias and Verdoliva, Luisa},
  booktitle={Proceedings of the IEEE/CVF Conference on Computer Vision and Pattern Recognition},
  pages={4356--4366},
  year={2024}
}

@article{li2025artificial,
  title={Is artificial intelligence generated image detection a solved problem?},
  author={Li, Ziqiang and Yan, Jiazhen and He, Ziwen and Zeng, Kai and Jiang, Weiwei and Xiong, Lizhi and Fu, Zhangjie},
  journal={arXiv preprint arXiv:2505.12335},
  year={2025}
}

@article{yang2025all,
  title={All patches matter, more patches better: Enhance ai-generated image detection via panoptic patch learning},
  author={Yang, Zheng and Chen, Ruoxin and Yan, Zhiyuan and Zhang, Ke-Yue and Fu, Xinghe and Wu, Shuang and Shu, Xiujun and Yao, Taiping and Ding, Shouhong and Qin, Zequn and others},
  journal={arXiv preprint arXiv:2504.01396},
  year={2025}
}

@inproceedings{cazenavette2024fakeinversion,
  title={Fakeinversion: Learning to detect images from unseen text-to-image models by inverting stable diffusion},
  author={Cazenavette, George and Sud, Avneesh and Leung, Thomas and Usman, Ben},
  booktitle={Proceedings of the IEEE/CVF Conference on Computer Vision and Pattern Recognition},
  pages={10759--10769},
  year={2024}
}

@article{ma2023exposing,
  title={Exposing the fake: Effective diffusion-generated images detection},
  author={Ma, Ruipeng and Duan, Jinhao and Kong, Fei and Shi, Xiaoshuang and Xu, Kaidi},
  journal={arXiv preprint arXiv:2307.06272},
  year={2023}
}

@inproceedings{fu2024iisan,
  title={IISAN: Efficiently adapting multimodal representation for sequential recommendation with decoupled PEFT},
  author={Fu, Junchen and Ge, Xuri and Xin, Xin and Karatzoglou, Alexandros and Arapakis, Ioannis and Wang, Jie and Jose, Joemon M},
  booktitle={Proceedings of the ACM SIGIR Conference on Research and Development in Information Retrieval},
  pages={687--697},
  year={2024}
}

@inproceedings{fang2025forensic,
  title={Forensic-moe: Exploring comprehensive synthetic image detection traces with mixture of experts},
  author={Fang, Mingqi and Li, Ziguang and Yu, Lingyun and Yang, Quanwei and Xie, Hongtao and Zhang, Yongdong},
  booktitle={Proceedings of the IEEE/CVF International Conference on Computer Vision},
  pages={17772--17782},
  year={2025},
  organization={IEEE}
}

@article{yan2026dual,
  title={Dual Frequency Branch Framework With Reconstructed Sliding Windows Attention for AI-Generated Image Detection},
  author={Yan, Jiazhen and Li, Ziqiang and Wang, Fan and He, Ziwen and Fu, Zhangjie},
  journal={IEEE Transactions on Information Forensics and Security},
  volume={21},
  pages={2313--2325},
  year={2026},
  publisher={IEEE}
}

@article{guan2025noise,
  title={Noise-informed diffusion-generated image detection with anomaly attention},
  author={Guan, Weinan and Wang, Wei and Peng, Bo and He, Ziwen and Dong, Jing and Cheng, Haonan},
  journal={IEEE Transactions on Information Forensics and Security},
  volume={20},
  pages={5256--5268},
  year={2025},
  publisher={IEEE}
}

@article{zhang2026falcon,
  title={FALCON-Net: Feature Aggregation of Local Patterns for AI-Generated Image Detection},
  author={Zhang, Dengyong and Hu, Miao and Chen, Jiaxin and Chen, Changsheng and Wang, Jin and Song, Yun and Yang, Gaobo and Liao, Xin and Ding, Xiangling},
  journal={IEEE Transactions on Information Forensics and Security},
  volume={21},
  pages={3842--3855},
  year={2026},
  publisher={IEEE}
}

@article{deng2026clipada,
  title={CLIP-ADA: CLIP-Guided Artifact-Invariant Generalizable Synthetic Image Detection},
  author={Deng, Jingyi and Zhang, Chong and Xu, Chenken and Lin, Chenhao and Zhao, Zhengyu and Liu, Shuai and Wang, Qian and Shen, Chao},
  journal={IEEE Transactions on Information Forensics and Security},
  volume={21},
  pages={3588--3601},
  year={2026},
  publisher={IEEE}
}

@article{tang2025extensible,
  title={Towards extensible detection of AI-generated images via content-agnostic adapter-based category-aware incremental learning},
  author={Tang, Shuai and He, Peisong and Li, Haoliang and Wang, Wei and Jiang, Xinghao and Zhao, Yao},
  journal={IEEE Transactions on Information Forensics and Security},
  volume={20},
  pages={2883--2898},
  year={2025},
  publisher={IEEE}
}

@inproceedings{koutlis2024leveraging,
  title={Leveraging Representations from Intermediate Encoder-Blocks for Synthetic Image Detection},
  author={Koutlis, Christos and Papadopoulos, Symeon},
  booktitle={European Conference on Computer Vision},
  pages={394--411},
  year={2024},
  organization={Springer}
}

@article{qin2025scaling,
  title={Scaling Up {AI}-Generated Image Detection via Generator-Aware Prototypes},
  author={Qin, Ziheng and Ji, Yuheng and Tao, Renshuai and Tian, Yuxuan and Liu, Yuyang and Wang, Yipu and Zheng, Xiaolong},
  journal={arXiv preprint arXiv:2512.12982},
  year={2025}
}
\bibliographystyle{IEEEtran}

\end{document}